\documentclass[11pt]{article}

\usepackage[preprint]{acl}

\usepackage{times}
\usepackage{latexsym}
\usepackage{amsmath}
\usepackage{multirow}

\usepackage[T1]{fontenc}

\usepackage[utf8]{inputenc}

\usepackage{microtype}

\usepackage{inconsolata}

\usepackage{graphicx}
\usepackage{booktabs}
\usepackage{amssymb}

\newcommand{\benchmark}{\texttt{TimeVista}}

\newcommand{\TimeVistaNumSamples}{5,563}
\newcommand{\TimeVistaNumSamplesReal}{5,163}
\newcommand{\TimeVistaNumSamplesSynthetic}{400}

\usepackage{tcolorbox}

\definecolor{myboxcolor}{RGB}{245,245,245} 
\definecolor{myframe}{RGB}{0,0,128} 

\newtcolorbox{mybody}{
  colback=myboxcolor,
  colframe=myframe,
  boxrule=1pt, 
  left=1pt,
  right=1pt,
  top=1pt,
  bottom=1pt,
}

\title{TimeVista: Exploring and Exploiting Vision-Language Models\\ as Judges for Time Series Forecasting}

\author{
 \textbf{Zhi Chen}\textsuperscript{$*$},
  \textbf{Yuxuan Wang}\textsuperscript{$*$},
  \textbf{Jialong Wu}\textsuperscript{$\dagger$},
 \textbf{Yong Liu},
\\
 \textbf{Haoran Zhang},
 \textbf{Xingjian Su},
 \textbf{Jianmin Wang},
 \textbf{Mingsheng Long}\textsuperscript{$\dagger$} \\
 \textsuperscript{1}School of Software, BNRist, Tsinghua University, Beijing 100084, China,
\\
  {\tt\small chenzhi21@mails.tsinghua.edu.cn, wangyuxu22@mails.tsinghua.edu.cn} \\
{\tt\small wujialong0229@gmail.com, mingsheng@tsinghua.edu.cn}
}

\begin{document}
\maketitle

\let\thefootnote\relax\footnotetext{$^*$Equal contribution.}
\let\thefootnote\relax\footnotetext{$^\dagger$Corresponding author.}

\begin{abstract}
High-quality time series forecasting is pivotal for real-world decision-making. However, traditional point-wise metrics often fail to reveal complex temporal patterns and align poorly with human intuitive preferences. While the ``LLM-as-a-Judge'' paradigm has revolutionized text evaluation by providing flexible, human-aligned judgment, its application to time series remains largely unexplored. In this paper, we leverage Vision-Language Models (VLMs) as judges for time series forecasting, harnessing their ability to comprehend time series plots grounded in textual information. Specifically, we propose a novel framework integrating micro- and macro-level judgments informed by contextual information to evaluate time series forecasting. To this end, we introduce \texttt{TimeVista}, a comprehensive VLM-as-a-Judge benchmark comprising \TimeVistaNumSamples{} time series samples paired with detailed evaluation rubrics. Extensive meta-evaluations demonstrate that VLMs are highly reliable judges, achieving significantly higher consistency with human preferences than conventional metrics. Building upon our benchmark, we comprehensively assess recent Time Series Foundation Models (TSFMs) under the VLM-as-a-Judge paradigm. Our results demonstrate that VLMs serve as robust and interpretable judges, providing a comprehensive, human-aligned standard for evaluating time series models.
\end{abstract}

\section{Introduction}

\begin{figure}[t]
  \includegraphics[width=\columnwidth]{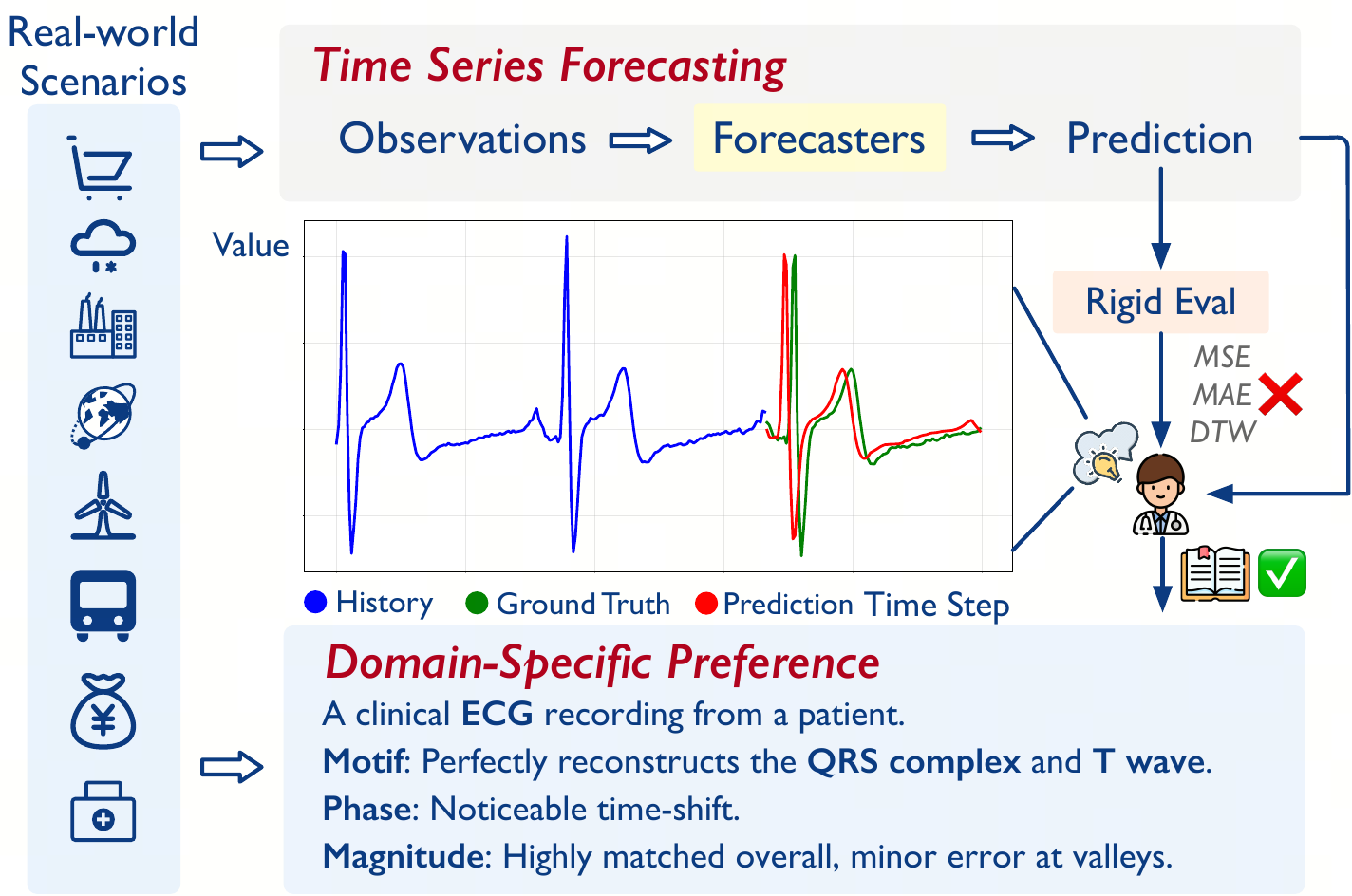}
  \caption{Traditional point-wise metrics fail to reflect domain-specific preferences in real-world scenarios.}
  \label{fig:intro}
  \vspace{-15pt}
\end{figure}

Time series forecasting plays a foundational role in decision-making across diverse real-world applications, ranging from financial market analysis~\cite{tsay2005analysis} to weather prediction~\cite{wu2023interpretable}. While forecasting models have rapidly evolved from classical statistical methods to Time Series Foundation Models (TSFMs)~\cite{liang2024foundation, miller2024survey} and multimodal architectures~\cite{wu2025aurora, ahamed2026reasoning}, the evaluation paradigm remains stagnant, relying on point-wise metrics, notably mean squared error (MSE) or mean absolute error (MAE).

This creates a fundamental mismatch: while the practical utility of a forecast hinges on structural fidelity and domain context, standard metrics fail to convey such rich information. For example, in Electrocardiogram (ECG) forecasting shown in Figure~\ref{fig:intro}, preserving waveform morphology is medically critical, whereas exact temporal alignment is secondary. Yet, operating in a numerical vacuum, standard point-wise metrics often penalize useful shifted peaks more heavily than useless smoothed waves, directly contradicting human intuition by prioritizing exact numerical precision over meaningful structural alignment.

Similar evaluation dilemmas exist in Natural Language Generation~\cite{gu2024survey}, where traditional n-gram-based metrics like BLEU~\cite{papineni2002bleu} and ROUGE~\cite{lin2004rouge} fail to reflect human preferences, prompting the rise of the ``LLM-as-a-Judge'' for more human-aligned feedback~\cite{gu2024survey, chen2024mllm}. This success motivates us to leverage the understanding capabilities of LLMs for time series to address the misalignment between evaluation metrics and human preferences. However, extending text-centric LLMs to this domain is hindered by severe modality gaps. First, discrete tokenization conflicts with the continuous numerical data~\cite{jin2024time}, impairing magnitude comprehension. Second, extensive time series sequences expose LLMs to severe attention degradation, known as the ``lost in the middle'' effect~\cite{liu2024lost}. Third, representing time series as flat text obscures essential temporal patterns, such as trends and phases.

To overcome the modality limitations of text-only LLMs, we draw inspiration from human experts who intuitively evaluate forecasting quality by inspecting visual plots, introducing the Vision-Language Model (VLM)-as-a-Judge paradigm to time series.
Unlike general image or chart understanding, interpreting time series plots requires joint comprehension of temporal dynamics and domain context.
Therefore, we formulate the forecast evaluation as a two-level multimodal task. At the micro-level, VLMs assess fine-grained temporal patterns, including trend, phase, motif, and magnitude, directly from visualized curves. At the macro-level, the evaluation is grounded in domain knowledge to judge whether the forecasts align with domain-specific constraints and preferences.

To this end, we introduce \texttt{TimeVista}, the first VLM-as-a-Judge benchmark for time series forecasting, as illustrated in Figure \ref{fig:contribution}, containing \TimeVistaNumSamples{} time series samples paired with detailed evaluation rubrics. Through extensive meta-evaluations, we validate that VLMs are highly reliable judges with superior human alignment compared to conventional metrics. Further, we comprehensively assess 13 representative forecasting models, applying micro-level criteria to the existing GIFT-Eval benchmark~\cite{aksu2024gift} and utilizing \texttt{TimeVista} for macro-level evaluation. By shifting focus from rigid numerical precision to structural fidelity and practical utility, this work establishes a more human-aligned evaluation paradigm for time series forecasting. Our contributions are summarized as:
\begin{figure}[t]
    \centering
    \includegraphics[width=0.85\linewidth]{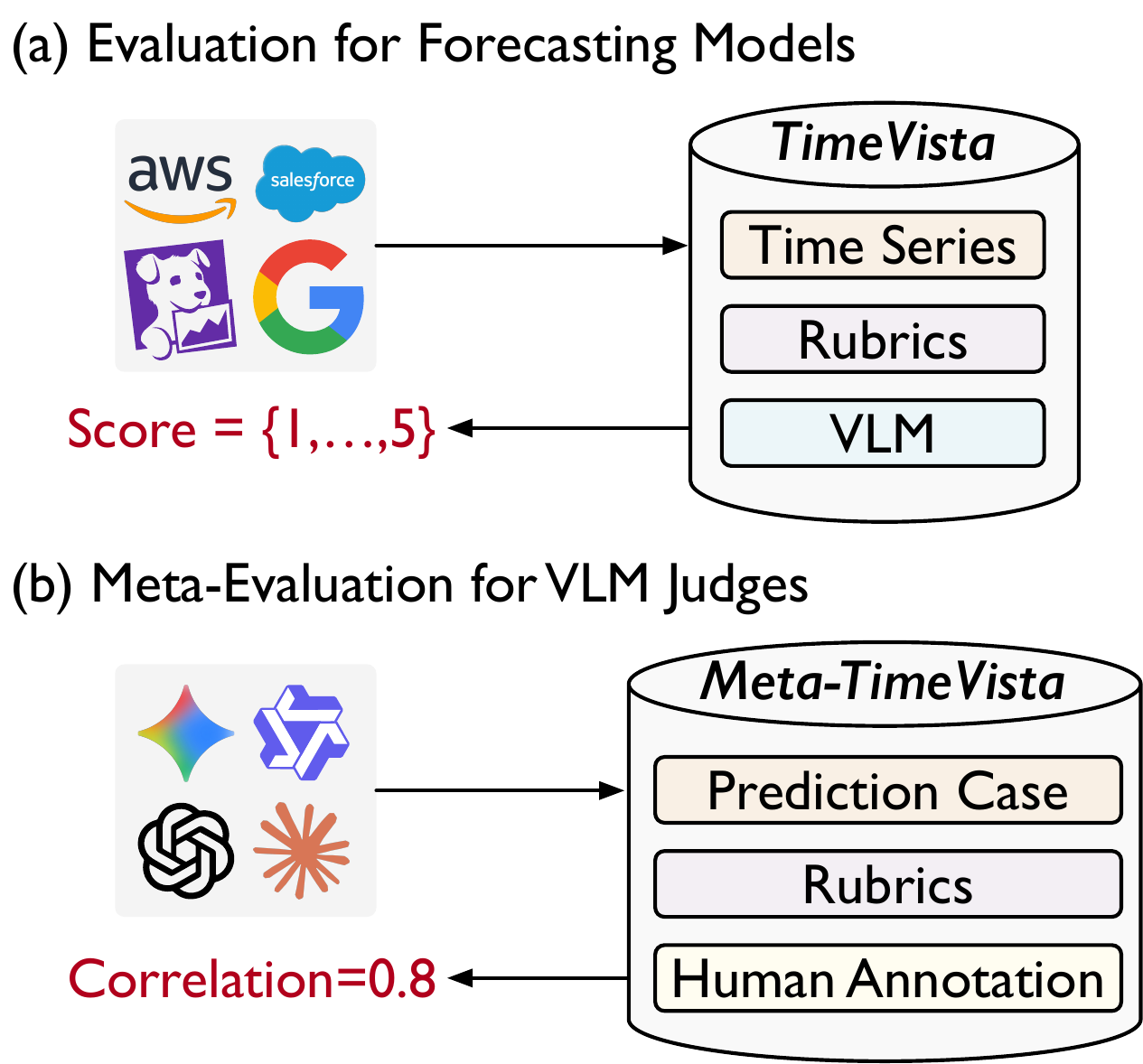}
    \caption{Illustration of the benchmarks proposed in this work. (a) Forecasting model evaluation on \texttt{TimeVista} using VLM judges with scoring rubrics. (b) Meta-evaluation on \texttt{Meta-TimeVista} to assess the alignment between VLM judges and human annotations.}
    \label{fig:contribution}
    \vspace{-10pt}
\end{figure}
\begin{itemize}
\item We explore and exploit the use of VLMs as judges for time series forecasting and introduce \texttt{TimeVista}, the first VLM-as-a-Judge benchmark, featuring \TimeVistaNumSamples{} time series paired with detailed evaluation rubrics.
\item We construct \texttt{Meta-TimeVista}, a meta-evaluation benchmark comprising 1,025 prediction cases generated from diverse models, each paired with human annotations.
\item Through extensive meta-evaluation on \texttt{Meta-TimeVista}, we demonstrate that VLM judges exhibit superior reliability and strong alignment with human preferences.
\item We conduct comprehensive evaluations of 13 representative forecasting models exploiting the VLM-as-a-Judge paradigm, leveraging GIFT-Eval and \texttt{TimeVista} for micro- and macro-level assessments respectively.
\end{itemize}

\section{Related Work}

\subsection{Time Series Forecasting Benchmarks}


Evaluation benchmarks have played a central role in the development of time series forecasting. Previous deep time series forecasters were primarily evaluated on dataset-specific benchmarks such as ETT and Weather~\cite{wang2026deep}, with evaluation largely based on point-wise numerical errors.
Recently, the emergence of Time Series Foundation Models (TSFMs)~\cite{das2023decoder, woo2024unified, ansari2025chronos} shifted the community's focus toward cross-domain generalization and zero-shot forecasting. To accommodate this paradigm shift, several large-scale benchmarks have emerged. For instance, GIFT-Eval~\cite{aksu2024gift} compiles a massive collection of diverse datasets to evaluate the universal zero-shot forecasting capabilities. Similarly, fev-bench~\cite{shchur2025fev} introduces a multivariate forecasting benchmark, while recent TIME~\cite{qiao2026s} introduces a task-centric benchmark and proposes a novel pattern-level evaluation perspective. 

\begin{figure}[t]
  \includegraphics[width=\linewidth]{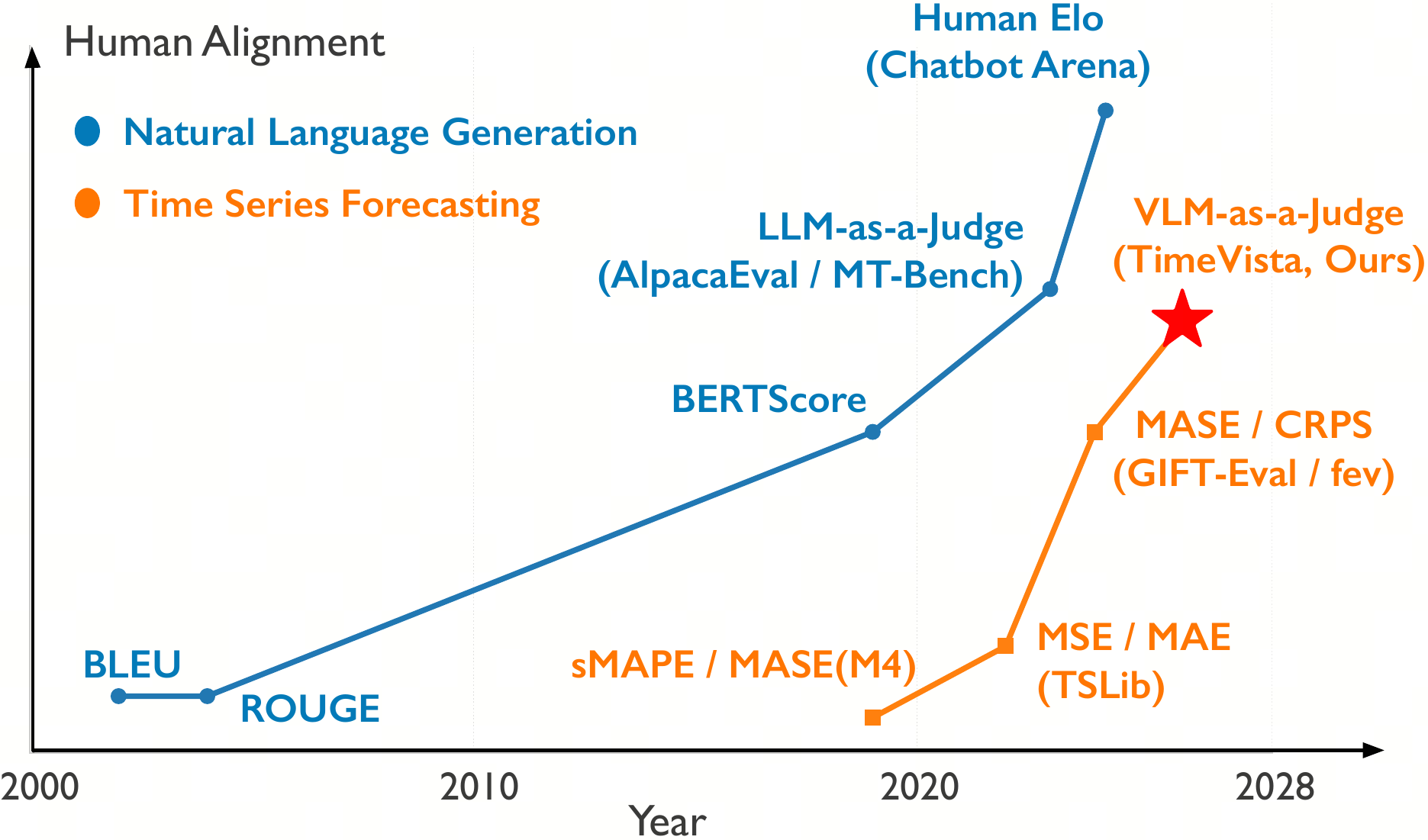}
  \vspace{-20pt}
  \caption{The paradigm shift of evaluation benchmarks across natural language generation and time series forecasting from rigid metrics toward model-based judging.}
  \label{fig:benchmark_evol}
  \vspace{-15pt}
\end{figure}

While these efforts substantially broaden data coverage and evaluation protocols, most benchmarks still focus on purely numerical, point-wise errors. To bridge this, as in Figure~\ref{fig:benchmark_evol}, \texttt{TimeVista} leverages the VLM-as-a-Judge paradigm to evaluate predictions from visual plots grounded in textual contexts, providing a multimodal and human-aligned complement to existing metrics.

\subsection{LLM-as-a-Judge Evaluation}
In the field of Natural Language Processing (NLP)~\cite{devlin2019bert, zhao2023survey}, evaluating open-ended generation has long suffered from the limitations of lexical metrics like BLEU and ROUGE \cite{papineni2002bleu, lin2004rouge, zhang2019bertscore}. Relying solely on rigid n-gram overlap, these metrics fail to capture semantic equivalence, structural coherence, and human preferences. To address this, the ``LLM-as-a-Judge'' paradigm emerged as a revolutionary alternative \cite{gu2024survey}, where strong LLMs are prompted to produce rubric-guided assessments and explanations that often correlate better with human preferences~\cite{zheng2023judging,chiang2024chatbot}. This paradigm has been widely adopted in evaluating dialogue systems, instruction following, and reasoning tasks \cite{bai2023benchmarking, fu2024gptscore}. More recently, this approach has been extended to the multimodal domain, giving rise to Vision-Language Models (VLMs) as judges for assessing image-text alignment and visual reasoning capabilities \cite{lee2024prometheus, lu2024mathvista}.
Despite its success in bypassing rigid metrics to provide human-aligned feedback, this paradigm remains unexplored in time series analysis. Inspired by how human experts evaluate prediction quality through visual inspection, the proposed \texttt{TimeVista} leverages VLMs to bypass these textual constraints. By rendering forecasts as visual plots grounded in textual context, we empower VLMs to jointly assess structural dynamics and semantic constraints, yielding holistic, human-aligned assessments.

\section{TimeVista}

\subsection{Problem Formulation}

In this paper, we focus on the evaluation of time series forecasting. Given a historical lookback window $\mathbf{x} \in \mathbb{R}^L$, a forecasting model $\mathcal{F}_{\theta}$ generates a prediction ${\hat{\mathbf{y}}} \in \mathbb{R}^T$ over a future horizon $T$, where ${\hat{\mathbf{y}}} = \mathcal{F}_{\theta}(\mathbf{x})$. Let $\mathbf{y} \in \mathbb{R}^T$ be the corresponding ground-truth.
Following LLM-as-a-Judge \cite{gu2024survey}, we cast forecasting evaluation as a context-aware multimodal judgment task. Let $\mathbf{V}$ denote the high-resolution plot rendering the historical input $\mathbf{x}$, the ground truth $\mathbf{y}$, and the prediction ${\hat{\mathbf{y}}}$ in contrasting colors, and $\mathcal{C}$ be the textual context specifying the scoring rubric. The VLM-as-a-Judge paradigm is formulated as:
\begin{equation}
    s, \mathcal{R} \leftarrow \mathcal{P}_{\mathrm{VLM}}(\mathbf{V}, \mathcal{C}),
\end{equation}
where $s\in\{1, \dots, 5\}$ is the Likert score, and $\mathcal{R}$ is the generated rationale, ensuring interpretability.


\subsection{Evaluation Protocol}
\label{sec:protocol}
The semantic information of a time series is embedded within its temporal variations~\cite{wu2022timesnet}, spanning both fine-grained temporal patterns and global dynamics. Building on this insight, we propose a two-level evaluation protocol. Specifically, the micro-level evaluation measures the fidelity of structural temporal patterns, whereas the macro-level evaluation verifies whether global dynamics comply with real-world operational constraints. 





\begin{figure*}[t]
  \includegraphics[width=\textwidth]{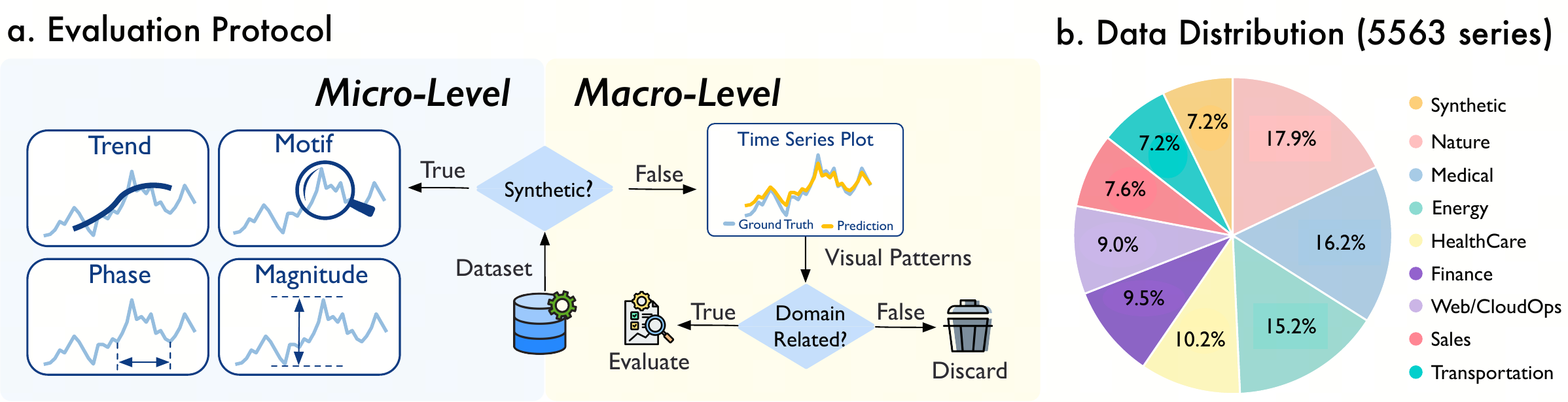}
  \caption{Evaluation protocol and data distribution of the proposed TimeVista. \textbf{(a)} The proposed protocol integrates  micro- and macro-level judgments. \textbf{(b)} The distribution of the benchmark dataset across diverse real-world domains.}
  \label{fig:main}
  \vspace{-10pt}
\end{figure*}

\subsubsection{Micro-Level} 
\label{micro_feature}

Unlike generic statistical charts, time series plots depict the continuous change of variables over time. To ensure a reliable and effective evaluation, it is essential to parse the underlying structural features directly from the visualized temporal sequences. In this section, we identify four key components that comprehensively characterize time series dynamics. To this end, we draw inspiration from classical time series decomposition, and mathematically formulate a series $y(t)$ as a combination of distinct structural components:
\begin{equation}
\label{time_series_decomposition}
y(t) = \mathcal{T}(t) + \mathcal{A}(t) \cdot \mathcal{M}\big( \mathcal{P}(t) \big)
\end{equation}
This formulation motivates our four micro-level evaluation dimensions, decoupling the curve's local geometry from global scale and temporal shifts:

\begin{itemize}
    \item \textbf{Trend ($\mathcal{T}$):} The global low-frequency trajectory, capturing overall directional movement.
    \item \textbf{Magnitude ($\mathcal{A}$):} The vertical amplitude, representing the scale of temporal variations.
    \item \textbf{Motif ($\mathcal{M}$):} The local geometric shape, representing recurring patterns and sub-series.
    \item \textbf{Phase ($\mathcal{P}$):} The horizontal timing, capturing the alignment of peaks and troughs.
\end{itemize}
These distinct yet complementary dimensions together enable a fine-grained and comprehensive evaluation of the visualized series.

\subsubsection{Macro-Level}
Micro-level geometric fidelity alone does not guarantee real-world practical utility, as the tolerability of prediction behaviors is highly context-sensitive. A structural deviation benign in one domain may prove catastrophic in another, necessitating a contextualized evaluation protocol. While micro-level metrics focus on specific temporal patterns, human experts inherently evaluate predictions top-down, leveraging domain knowledge to gauge global viability. Inspired by this cognitive hierarchy, the macro-level evaluation introduces distinct rubrics explicitly formulated to align with domain-specific preferences and downstream decision-making.

\subsection{Dataset Construction}

\label{sec:dataset}

We construct \texttt{TimeVista}, a comprehensive benchmark comprising \TimeVistaNumSamples{} time series samples with meticulously crafted rubrics. It contains both synthetic data and real-world data to  evaluate micro-level and macro-level forecasting performance.

\paragraph{Time Series Collection} 

We construct our real-world subset by collecting and filtering data from established benchmarks, yielding a curated collection of \TimeVistaNumSamplesReal{} samples spanning eight diverse domains (Figure \ref{fig:main}). To ensure data quality and practical alignment, we remove missing values, discard irrelevant variables, and accommodate various sampling frequencies and input/output lengths. Finally, we conduct visual inspections to discard series lacking clear temporal patterns for robust evaluation.

While real-world data reflects practical complexity, its structural components are inherently intertwined and relatively limited. To overcome this, we construct a synthetic subset comprising \TimeVistaNumSamplesSynthetic{} time series samples by orthogonally controlling the four micro-level features defined in Section~\ref{micro_feature}. This controlled synthesis acts as a stress test; by varying one specific dimension while keeping others constant, we can cover rare corner cases and precisely diagnose a model's structural weaknesses without feature coupling. Detailed synthesis procedures are provided in Appendix~\ref{app:synthesis}. In total, \benchmark{} comprises \TimeVistaNumSamples{} series, with \TimeVistaNumSamplesReal{} from real-world domains and \TimeVistaNumSamplesSynthetic{} generated through synthesis.


\paragraph{Rubrics Construction} 
The cornerstone of \benchmark{} lies in its evaluation rubrics. At the micro-level, we formulate five-point Likert scales across trend, motif, phase, and magnitude, grounded in foundational time series theory~\cite{hamilton2020time}. Each criterion specifies explicit visual cues such as slope deviations and peak-valley alignment to ensure operational clarity. At the macro-level, we design domain-specific rubrics that translate visual patterns into practical utility under real-world constraints. These rubrics are then reviewed by human experts from corresponding domains. Through collaborative discussions, the experts iteratively refine the evaluation criteria to guarantee domain relevance, logical consistency, and clarity.

\paragraph{Standardized Visualization} 


We plot each forecasting instance, comprising the historical context, model prediction, and ground-truth as a 2D line chart with explicit numerical axis labels, aligned grids, high-contrast colors, and a clear legend. We strip non-essential aesthetic elements, such as decorative backgrounds, to minimize visual noise.

\section{Meta-Evaluation of VLM Judges}
\label{sec:meta_eval}
We conduct a rigorous meta-evaluation to explore this novel VLM-as-a-Judge paradigm, systematically validating its reliability and superiority.

\subsection{Meta-TimeVista Benchmark}

To systematically validate the VLM's capabilities, we construct \texttt{Meta-TimeVista} for meta-evaluation, including time series and their corresponding textual rubrics from the \texttt{TimeVista} dataset, and then employ eight TSFMs detailed in Section~\ref{sec:main_exp} to generate diverse zero-shot predictions. This process ultimately yields 1,025 distinct prediction cases, comprising 480 synthetic and 545 real-world cases.

To strictly implement the two-level evaluation protocol, this benchmark is strategically partitioned into two subsets based on data provenance. We use the synthetic subset to conduct micro-level evaluations across various temporal patterns, and the real-world subset for macro-level evaluations. Since a reliable macro-level evaluation hinges on the VLM's robust perception of the overall geometry, we introduce an intermediate \textit{macro-level-w/o-knowledge} variant to verify this prerequisite. For clarity, we denote the original macro-level evaluation as \textit{macro-level-w/-knowledge} in Section~\ref{sec:meta_eval}.

\paragraph{Human Annotation}

To establish a reliable ground truth, five time series experts independently evaluated all prediction cases using the same 1–5 Likert rubrics as the VLM judges, as provided in Appendix~\ref{app:rubrics}, assessing both micro- and macro-level performance. The final ground truth score for each case was determined by the mode of the ratings. To validate these annotations, we calculated the inter-annotator agreement (IAA). As shown in Table~\ref{tab:IAA_result}, the consensus across both levels, highlighted by Spearman's $\rho$ consistently exceeding $0.89$ across all dimensions, demonstrates that our evaluation task is well-defined, the rubrics are unambiguous, and the resulting ground truth is highly reliable for benchmarking VLM judges.

\subsection{Evaluation Setup}

\paragraph{Metrics}
To evaluate VLM-human alignment, we calculate Spearman's $\rho$ and Kendall's $\tau$ between paired VLM scores $X$ and human ground truth $Y$. Spearman's $\rho$ is computed as the Pearson correlation of rank-transformed scores to assess monotonic agreement, while Kendall's $\tau$ measures rank concordance based on the relative ordering of all sample pairs. Higher values indicate stronger alignment with human preference.

\begin{table*}[t]
\caption{Spearman ($\rho$) and Kendall-Tau ($\tau$) correlations between VLM judges and human annotations across micro- and macro-level evaluation. Overall, VLMs exhibit strong alignment with human preferences.}
\vspace{-5pt}
\setlength{\tabcolsep}{4.8pt}
\renewcommand{\arraystretch}{1.2} 
\small
\begin{tabular}{l|cc|cc|cc|cc|cc|cc}
\toprule
\multicolumn{1}{c|}{\textbf{Level}} & \multicolumn{8}{c|}{\textbf{Micro-Level}} & \multicolumn{4}{c}{\textbf{Macro-Level}} \\
\cmidrule(lr){2-11}\cmidrule(lr){12-13}
\multicolumn{1}{c|}{\multirow{2}[1]{*}{\textbf{Metrics}}} & \multicolumn{2}{c|}{\textbf{Trend}} & \multicolumn{2}{c|}{\textbf{Motif}} & \multicolumn{2}{c|}{\textbf{Phase}} & \multicolumn{2}{c|}{\textbf{Magnitude}} & \multicolumn{2}{c|}{\textbf{w/o knowledge}} & \multicolumn{2}{c}{\textbf{w/ knowledge}} \\
\cmidrule(lr){2-3} \cmidrule(lr){4-5} \cmidrule(lr){6-7}\cmidrule(lr){8-9} \cmidrule(lr){10-11} \cmidrule(lr){12-13}
& $\rho$ & $\tau$ & $\rho$ & $\tau$ & $\rho$ & $\tau$ & $\rho$ & $\tau$ & $\rho$ & $\tau$ & $\rho$ & $\tau$ \\
\toprule
Gemini-3.1-Pro & \scalebox{1.05}{\textbf{0.860}} & \scalebox{1.05}{\textbf{0.773}} & \scalebox{1.05}{\textbf{0.869}} & \scalebox{1.05}{\textbf{0.796}} & \scalebox{1.05}{0.831} & \scalebox{1.05}{0.764} & \scalebox{1.05}{\textbf{0.907}} & \scalebox{1.05}{\textbf{0.849}} & \scalebox{1.05}{\textbf{0.909}} & \scalebox{1.05}{\textbf{0.850}} & \scalebox{1.05}{\textbf{0.865}} & \scalebox{1.05}{\textbf{0.820}} \\
Qwen3.6-Plus & \scalebox{1.05}{0.847} & \scalebox{1.05}{0.759} & \scalebox{1.05}{0.854} & \scalebox{1.05}{0.772} & \scalebox{1.05}{0.768} & \scalebox{1.05}{0.690} & \scalebox{1.05}{0.870} & \scalebox{1.05}{0.805} & \scalebox{1.05}{0.888} & \scalebox{1.05}{0.824} & \scalebox{1.05}{0.859} & \scalebox{1.05}{0.806} \\
GPT-5.4 & \scalebox{1.05}{0.836} & \scalebox{1.05}{0.753} & \scalebox{1.05}{0.845} & \scalebox{1.05}{0.772} & \scalebox{1.05}{\textbf{0.854}} & \scalebox{1.05}{\textbf{0.781}} & \scalebox{1.05}{0.837} & \scalebox{1.05}{0.769} & \scalebox{1.05}{0.871} & \scalebox{1.05}{0.802} & \scalebox{1.05}{0.794} & \scalebox{1.05}{0.740} \\
Claude-Sonnet-4.6 & \scalebox{1.05}{0.807} & \scalebox{1.05}{0.728} & \scalebox{1.05}{0.802} & \scalebox{1.05}{0.724} & \scalebox{1.05}{0.690} & \scalebox{1.05}{0.609} & \scalebox{1.05}{0.826} & \scalebox{1.05}{0.755} & \scalebox{1.05}{0.853} & \scalebox{1.05}{0.786} & \scalebox{1.05}{0.792} & \scalebox{1.05}{0.732} \\
Seed-1.8 & \scalebox{1.05}{0.764} & \scalebox{1.05}{0.663} & \scalebox{1.05}{0.767} & \scalebox{1.05}{0.683} & \scalebox{1.05}{0.734} & \scalebox{1.05}{0.652} & \scalebox{1.05}{0.802} & \scalebox{1.05}{0.719} & \scalebox{1.05}{0.815} & \scalebox{1.05}{0.732} & \scalebox{1.05}{0.760} & \scalebox{1.05}{0.690} \\
LLaMa-4-Maverick & \scalebox{1.05}{0.749} & \scalebox{1.05}{0.667} & \scalebox{1.05}{0.728} & \scalebox{1.05}{0.644} & \scalebox{1.05}{0.704} & \scalebox{1.05}{0.629} & \scalebox{1.05}{0.763} & \scalebox{1.05}{0.680} & \scalebox{1.05}{0.770} & \scalebox{1.05}{0.694} & \scalebox{1.05}{0.674} & \scalebox{1.05}{0.608} \\
 \bottomrule
\end{tabular}
\label{tab:meta_alignment}
\vspace{-5pt}
\end{table*}

\paragraph{Evaluated Models}
Our experimental suite comprises a diverse array of state-of-the-art VLMs, including GPT-5.4~\cite{achiam2023gpt}, Claude-Sonnet-4.6, Gemini-3.1-Pro~\cite{team2023gemini}, LLaMA-4-Maverick~\cite{touvron2023llama}, Qwen-3.6-Plus~\cite{yang2025qwen3}, and Seed-1.8~\cite{seed2026seed1}. To ensure reproducibility, all model generations are configured with a temperature of $0$, enforcing deterministic outputs across all trials.

\subsection{Results}
\label{subsec:alignment}

\paragraph{Alignment Performance} Table \ref{tab:meta_alignment} presents the overall alignment results on both the micro-level and macro-level evaluations. The results demonstrate that advanced VLMs possess a remarkable alignment with human expert judgments. Leading models, particularly Gemini-3.1-Pro and Qwen-3.6-Plus, achieve exceptionally high correlation scores. More importantly, in the macro-level setting that demands contextual reasoning and a holistic perception of time series, open-source models like Qwen-3.6-Plus exhibit highly competitive performance. This validates the robustness of the VLM-as-a-Judge paradigm across diverse model families for both temporal pattern identification and domain-specific reasoning.

\begin{table}[h]
\centering 
\setlength{\tabcolsep}{9pt}
\renewcommand{\arraystretch}{1.2} 
\small
\caption{Comparison of Spearman ($\rho$) and Kendall-Tau ($\tau$) correlations of VLM judges and conventional metrics with human annotations on macro-level evaluation.}
\vspace{-5pt}
\begin{tabular}{l|cc|cc}
\toprule
\multicolumn{1}{c|}{\multirow{2}[1]{*}{\textbf{Metrics}}} & \multicolumn{2}{c|}{\textbf{w/o knowledge}} & \multicolumn{2}{c}{\textbf{w/ knowledge}} \\ 
\cmidrule(lr){2-3} \cmidrule(lr){4-5}
& $\rho$ & $\tau$ & $\rho$ & $\tau$ \\ 
\toprule
Random & -0.017 & -0.013 & 0.071 & 0.055 \\
$-$ \text{MSE} & 0.390 & 0.282 & -0.135 & -0.098 \\
$-$ \text{MAE} & 0.402 & 0.291 & -0.126 & -0.091 \\
$-$ \text{DTW} & 0.421 & 0.304 & -0.102 & -0.073 \\ \midrule
\text{VLM} & \textbf{0.909} & \textbf{0.850} & \textbf{0.865} & \textbf{0.820} \\
\bottomrule
\end{tabular}
\label{tab:numerical_cmp}
\vspace{-10pt}
\end{table}

\paragraph{Advantages over Traditional Metrics}
To validate the VLM-as-a-Judge paradigm, we compare its human alignment against a Random baseline and traditional metrics, including MSE, MAE, and DTW. Since these traditional metrics inherently measure error magnitude, we use their negative values to align with human scoring directions.
As shown in Table \ref{tab:numerical_cmp}, conventional metrics exhibit a pronounced misalignment with human experts, yielding marginal or even negative correlation scores, particularly in the macro-level-w/-knowledge evaluation. This discrepancy reveals that point-wise numerical errors are ill-equipped to capture the structural dynamics and domain-specific preferences required in practice. Conversely, by performing joint vision-language reasoning, the VLM judge consistently and substantially outperforms all baselines, highlighting the limitations of rigid metrics and supporting the shift toward more human-aligned evaluation.


\section{Analysis and Discussions}
\label{sec:analysis}

In this section, we provide an in-depth analysis of the VLM-as-a-Judge paradigm. Specifically, we investigate the necessity of visual representations and validate the robustness of the VLM judges.

\subsection{Efficacy of Visual Modality}

\begin{figure}[t]
  \centering
  \includegraphics[width=0.9\columnwidth]{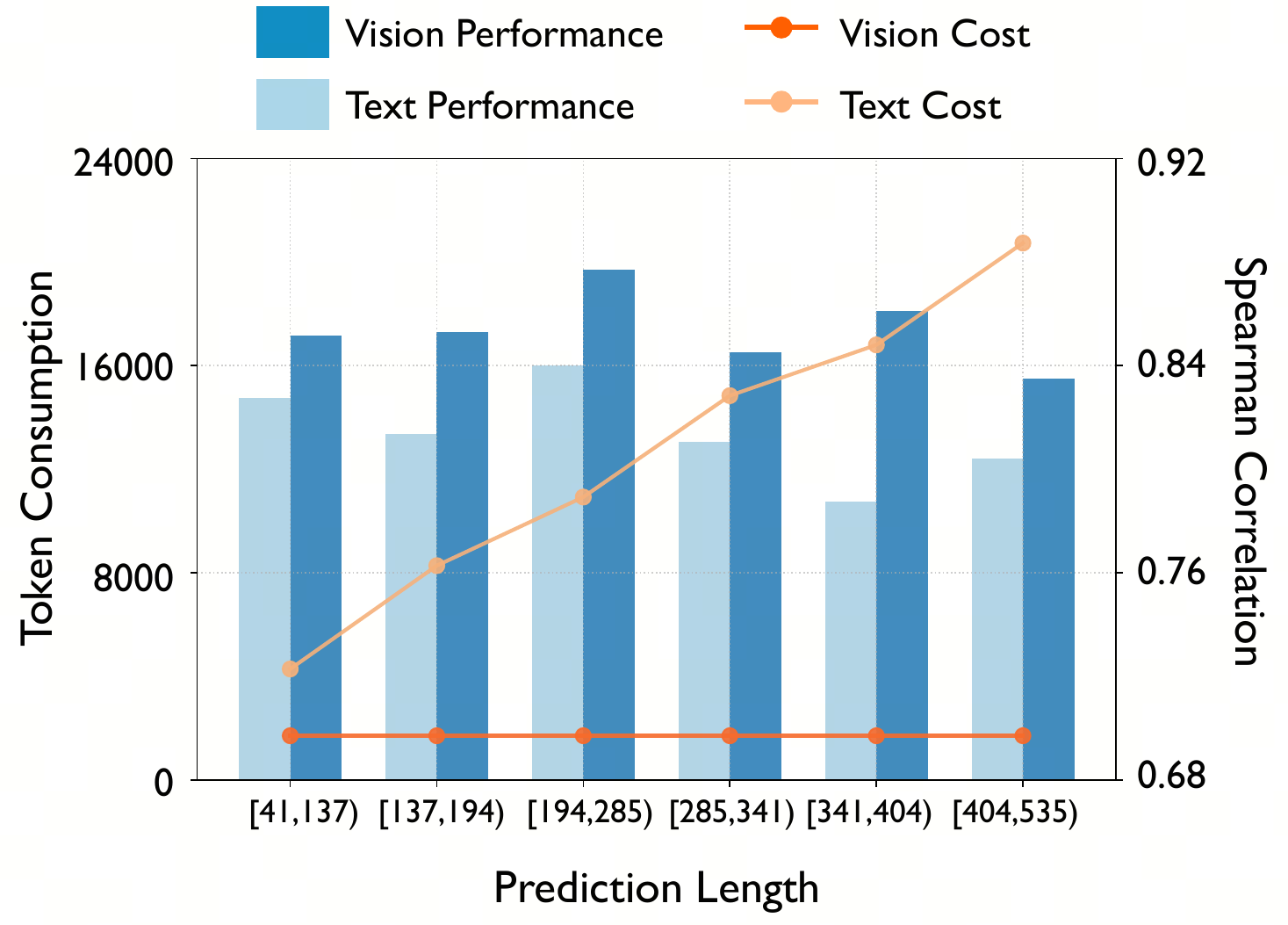}
  \vspace{-5pt}
  \caption{Comparison of vision-based and text-based evaluation across different prediction lengths.}
  \vspace{-15pt}
  \label{fig:text_vision}
\end{figure}

To justify the efficacy of representing time series as visual plots, we conduct a quantitative comparison between our visual-based evaluation and a text-based LLM-as-a-Judge baseline. Concretely, we implement the text baseline by replacing the visual plots with raw numerical text, using the same Gemini-3.1-Pro and keeping the prompts identical.
\paragraph{Computational Efficiency} As shown in Figure \ref{fig:text_vision}, representing numerical arrays as text imposes a token cost that scales linearly with the prediction length, rendering long-horizon forecasting prohibitively expensive. In contrast, visual rendering transforms the prediction into a single image, maintaining a constant token cost regardless of the prediction length. This compression drastically alleviates computational overhead while avoiding context overflow.

\paragraph{Judging Performance.} Beyond computational efficiency, visual evaluation significantly outperforms its text-based counterpart. Text tokenization inherently struggles with continuous numerical values, making text-only LLMs ill-equipped to comprehend long-horizon series. Visual rendering bypasses this bottleneck, enabling VLMs to capture temporal dynamics through overall visual perception. As illustrated in Figure \ref{fig:text_vision}, this visual approach consistently yields better correlation with human annotations across all prediction lengths.



\subsection{Robustness of VLM Judges}
A qualified visual judge must demonstrate robustness, showing resilience to non-semantic perturbations. To systematically validate this, we evaluate the VLM's sensitivity across two critical dimensions: visual styling and plot aspect ratios.

\begin{figure}[h]
  \centering
  \includegraphics[width=0.8\linewidth]{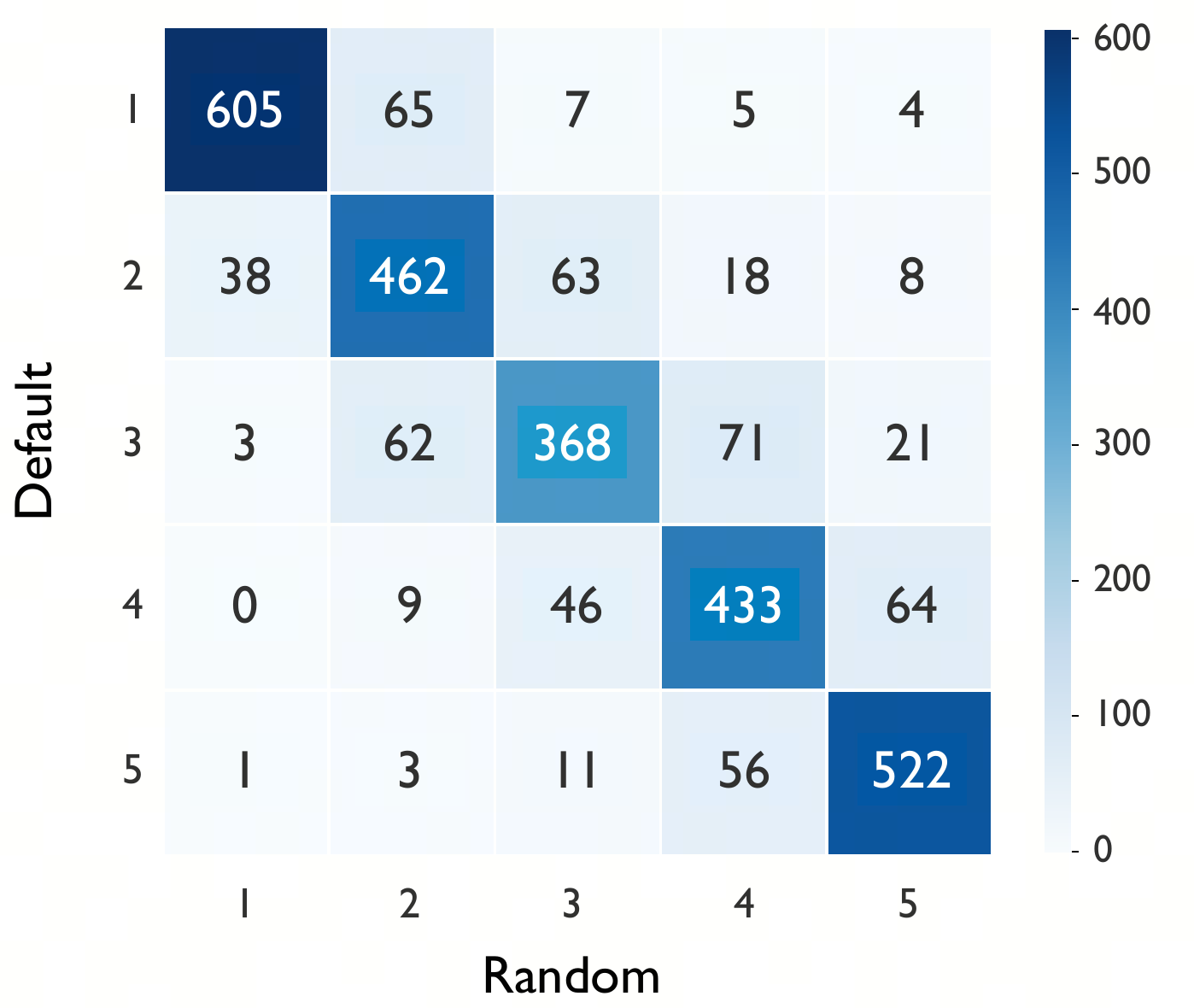}
  \vspace{-5pt}
  \caption{Sensitivity confusion matrix of VLM evaluation scores under default vs. randomized plot styles.}
  \vspace{-10pt}
  \label{fig:sensitivity}
\end{figure}

\paragraph{Visual Styling} We assess the VLM's robustness to visual styling by comparing its evaluation scores on default plots against randomized versions, where each sample is assigned a randomly selected color and line style. As shown in Figure~\ref{fig:sensitivity}, the VLM yields an 82.8\% exact match rate across these configurations. The highly diagonally dominant confusion matrix confirms that the VLM's judgment is not distracted by styling variations, but is consistently driven by the temporal patterns.

\paragraph{Aspect Ratios} We vary plot aspect ratios to confirm the VLM's robustness against geometric stretching or compression. As detailed in Table~\ref{tab:aspect_ratio_sensitivity}, the VLM judge remains highly stable across all configurations, with Spearman's $\rho$ and Kendall's $\tau$ fluctuating by less than 0.011, confirming robustness to non-semantic visual alterations.





\subsection{Qualitative examples of VLM Judgment}

\begin{figure}[h]
  \centering
  \includegraphics[width=\linewidth]{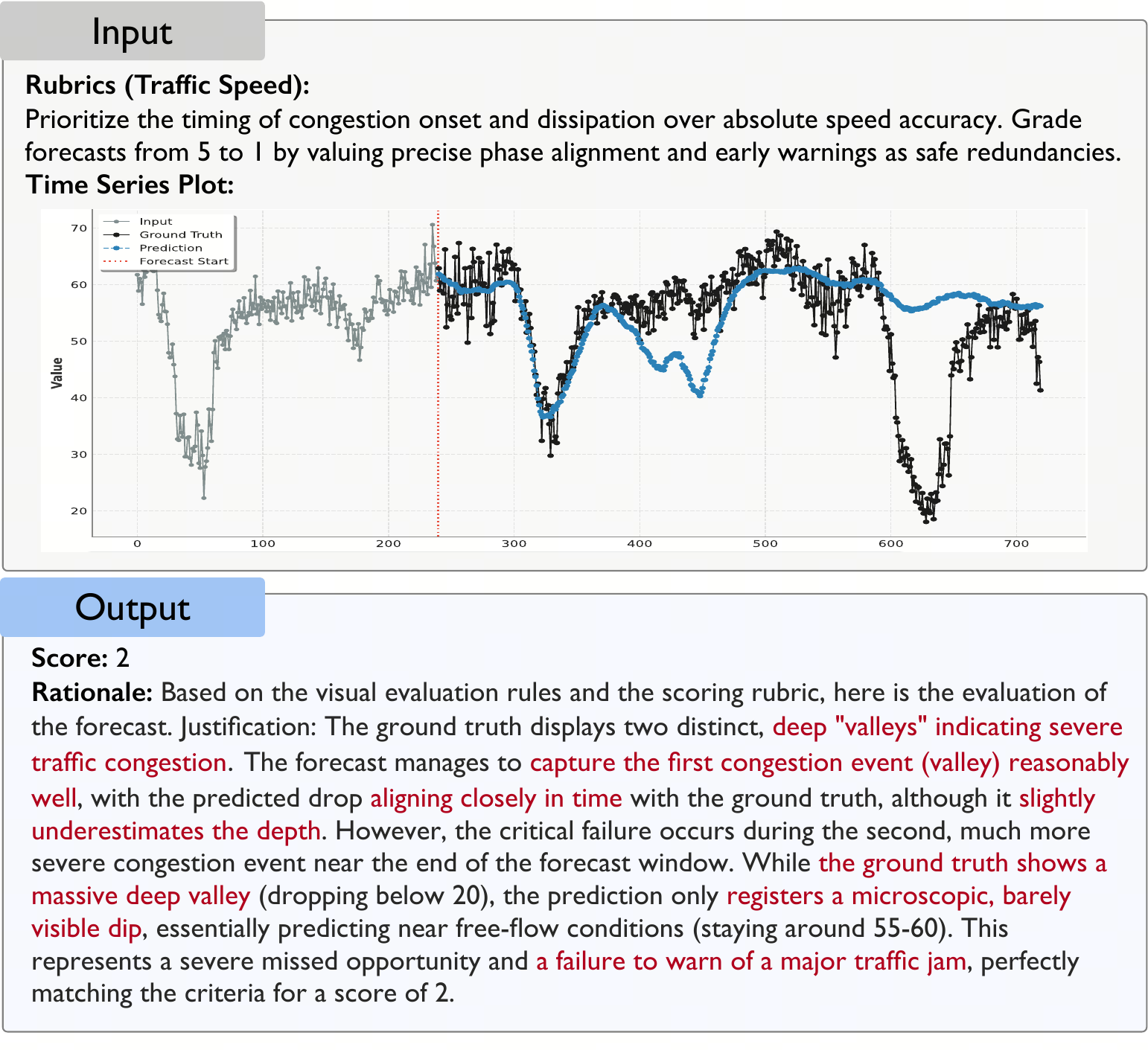}
  \vspace{-15pt}
  \caption{An example of judgment from Gemini-3.1-Pro for traffic speed prediction.}
  \vspace{-10pt}
  \label{fig:case_speed_2}
\end{figure}

To qualitatively demonstrate the reliability of the VLM judge, we provide a representative traffic speed forecasting case in Figure \ref{fig:case_speed_2}.
Grounded in domain knowledge, the VLM judge focuses exclusively on the two deep valleys in the ground truth, as sudden speed drops indicate critical traffic congestion, while ignoring prediction errors during high-speed, free-flow periods. It assigns a score of 2 with a rigorous justification: the forecasting model captures the first valley but completely fails to predict the second, much more severe congestion event. This highlights that the VLM judge goes beyond rigid point-wise metrics, successfully leveraging domain knowledge to visually target temporal patterns and deliver highly interpretable, human-aligned evaluations. We present more examples in Appendix~\ref{sec:app_example}.

\section{Evaluation of Time Series Models}

\subsection{Experiment Setup}
\label{sec:main_exp}
Based on our extensive meta-evaluation, we adopt the best-performing Gemini-3.1-Pro as the official evaluator. In this section, we fully exploit this VLM-as-a-Judge paradigm to comprehensively benchmark representative time series forecasting models. Specifically, we benchmark a diverse range of representative forecasting models, which can be categorized into three groups: 
(1) \textbf{Statistical methods}, including SMA~\cite{johnston1999sma}, AutoARIMA~\cite{shumway2017arima}, and AutoETS~\cite{petropoulos2022forecasting}; 
(2) \textbf{Time Series Foundation Models}, including Chronos~\cite{ansari2024chronos}, Sundial~\cite{liu2025sundial}, Timer-S1~\cite{liu2026timer}, Chronos2~\cite{ansari2025chronos}, TimesFM-2.5~\cite{das2023decoder}, TiRex~\cite{auer2026tirex}, Flowstate~\cite{graf2025flowstate}, and Moirai2~\cite{liu2025moirai}; 
and (3) \textbf{Multimodal models}, represented by Aurora~\cite{wu2025aurora}.


\begin{figure*}[t]
  \includegraphics[width=\linewidth
  ]{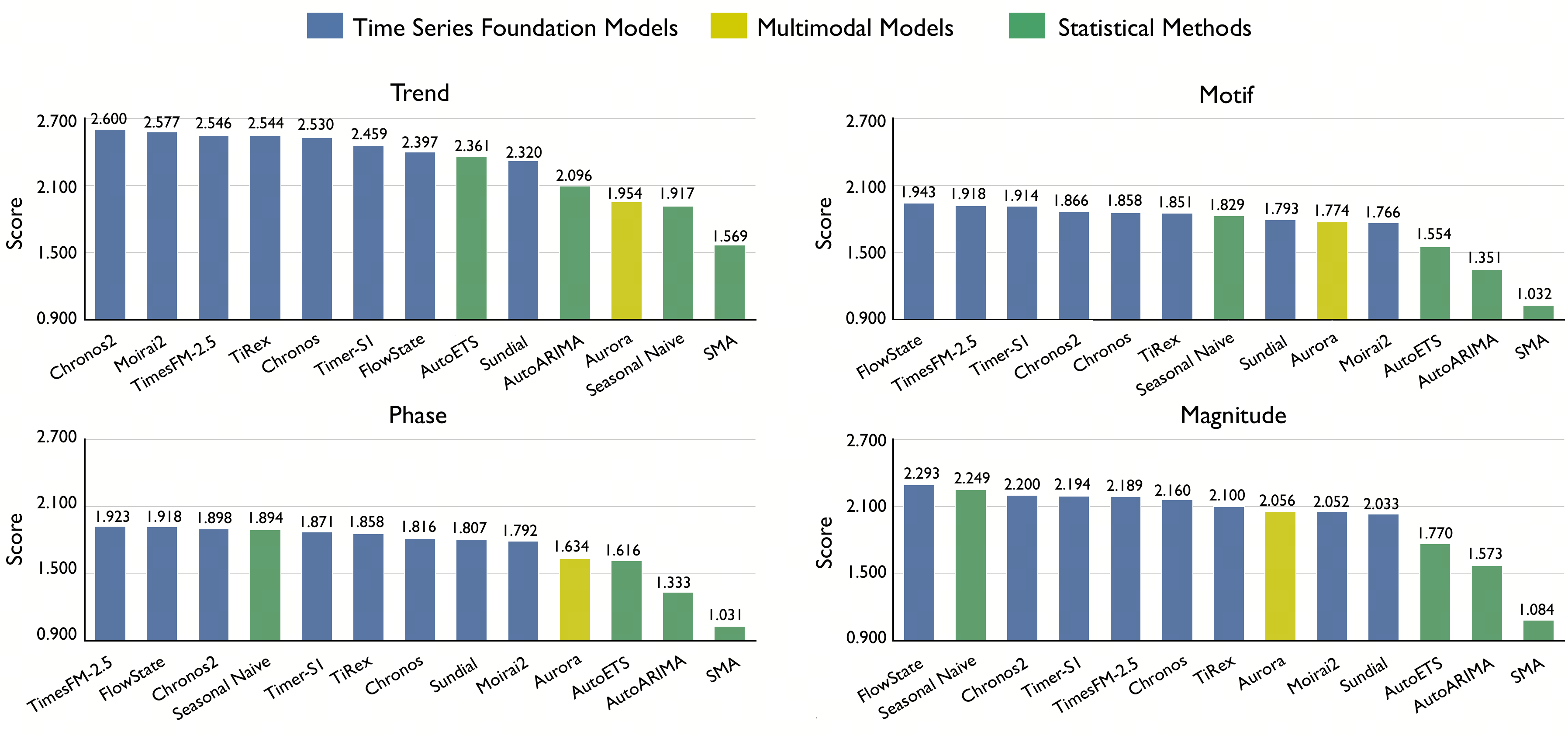}
  \vspace{-25pt}
  \caption{Micro-level evaluation of 13 representative forecasting models on GIFT-Eval benchmark.}
  \label{fig:gifteval_vista}
  \vspace{-10pt}
\end{figure*}

\subsection{Micro-Level Evaluation on GIFT-Eval}
To demonstrate the necessity of the VLM-as-a-Judge paradigm, we apply our micro-level protocol to re-evaluate forecasting models on the established GIFT-Eval benchmark~\cite{aksu2024gift}. Specifically, we randomly sample 846 time series and render their predictions into visual plots for VLM-based assessment across the four dimensions. As shown in Figure~\ref{fig:gifteval_vista}, model rankings vary significantly across these micro-level dimensions. This fine-grained evaluation offers a comprehensive view, revealing critical insights that traditional point-wise metrics fail to capture.
Specifically, in the trend dimension, Time Series Foundation Models, Chronos2~\cite{ansari2025chronos} and Moirai2~\cite{liu2025moirai}, achieve top performance, highlighting their superior capacity to capture long-range dependency. However, in the phase and magnitude dimensions, the simple Seasonal Naive baseline exhibits surprising competitiveness by preserving historical scales and periodicity. This contrast reveals a promising direction for TSFMs, namely, extending their capability to precise phase and magnitude alignment.


\subsection{Macro-Level Evaluation on TimeVista}
\begin{figure}[h]
  \includegraphics[width=0.95\linewidth]{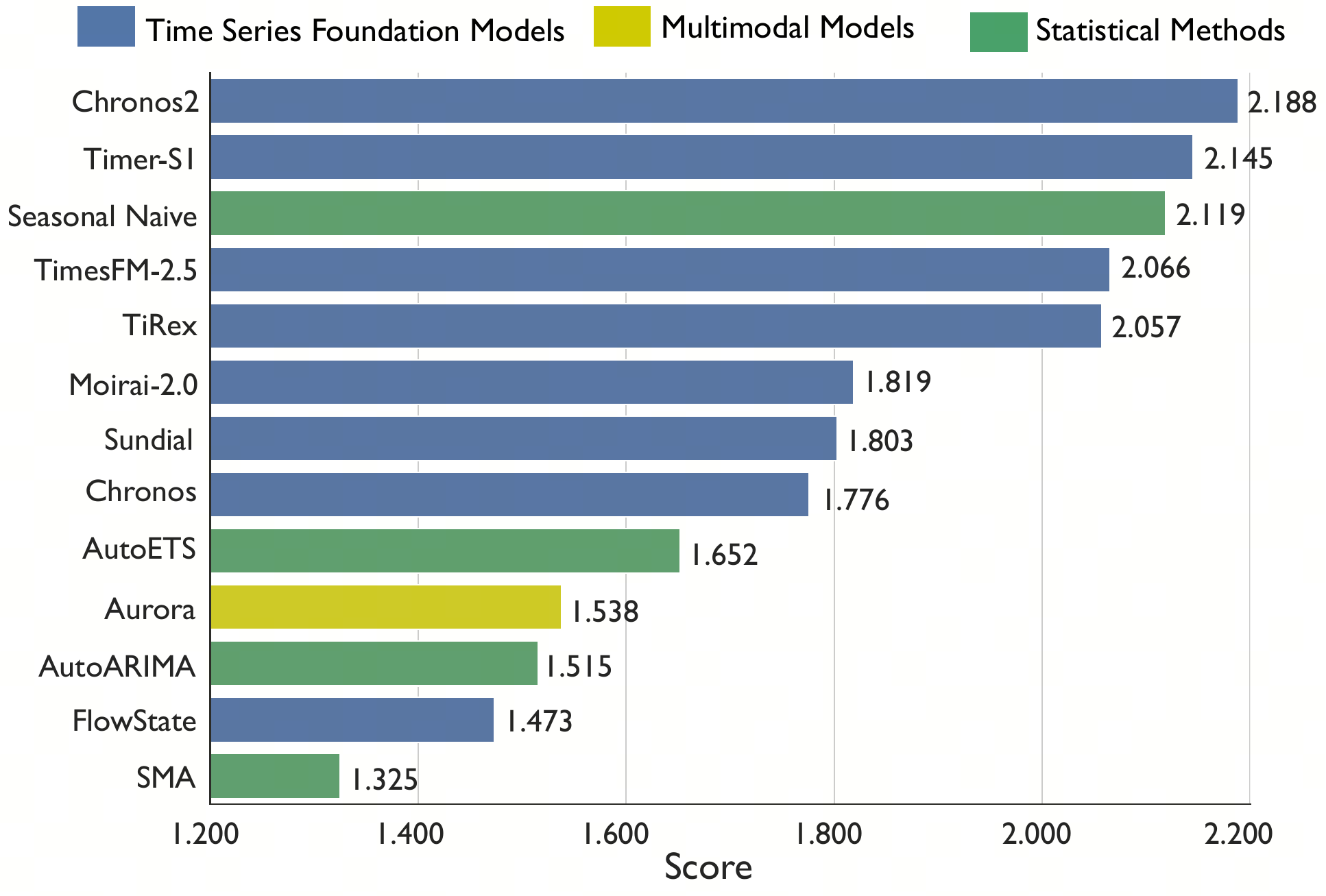}
  \vspace{-10pt}
  \caption{Macro-level evaluation of 13 representative forecasting models on \texttt{TimeVista} benchmark.}
  \label{fig:time_vista_benchmark}
  \vspace{-15pt}
\end{figure}

To conduct a macro-level evaluation, we randomly sample 945 time series uniformly across all domains from the \texttt{TimeVista} dataset. While current TSFMs excel in point-wise metrics~\cite{aksu2024gift}, they perform poorly under real-world domain-specific constraints as shown in Figure~\ref{fig:time_vista_benchmark}. Even the top-performing model Chronos2, achieves a modest score of only 2.188/5. 
This gap highlights a substantial mismatch between point-wise error minimization and practical utility, largely because standard metrics like MSE fail to reflect domain preferences. For instance, in energy domains, several advanced TSFMs frequently predict physically impossible negative production during valley periods. While MSE treats these critical physical violations as ordinary numerical deviations, the VLM judge successfully uncovers such anomalies by strictly enforcing domain-specific constraints.

Notably, the Seasonal Naive baseline ranks third overall, highlighting that preserving robust temporal patterns remains a cornerstone for practical deployment. Furthermore, the multimodal model Aurora significantly underperforms. Although capable of ingesting textual domain knowledge at the input, Aurora fails to process this textual information into practical constraints in its numerical forecasting process, leading to frequent violations of domain constraints. These results collectively underscore the value of \texttt{TimeVista} in uncovering critical insights for evaluating forecasting models that traditional evaluations overlook.

\section{Conclusion}
In this paper, we introduce \texttt{TimeVista}, a VLM-as-a-Judge benchmark to transcend point-wise metrics for time series predictions. By integrating visual plots with textual rubrics, our framework evaluates forecasting models across both micro-level patterns and macro-level domain utility. Extensive meta-evaluations demonstrate that VLMs serve as highly reliable, interpretable, and human-aligned judges.
Crucially, our evaluation reveals that state-of-the-art TSFMs still struggle with temporal phase alignment, exposing a critical blind spot of traditional metrics in capturing such pivotal dynamic features.
By introducing a novel perspective of evaluation beyond pure numerical precision, our work offers a vital feedback signal to guide the development of next-generation forecasting models.


\section*{Limitations}
Although \texttt{TimeVista} lays a solid foundation for evaluating time series forecasting under the VLM-as-a-Judge paradigm, demonstrating significant alignment with human preferences and providing interpretable evaluations, we acknowledge several limitations that point to promising avenues for future work.
The visual-based evaluation paradigm is constrained by the rendering resolution of plots and the visual acuity of VLMs. For extremely long-horizon or high-frequency sequences, visual clutter may lead to information loss, making it difficult for VLMs to discern fine-grained geometric details. Moreover, the proposed evaluation method currently relies on proprietary VLMs, which introduces API costs and potential reproducibility issues. To address these issues, developing and fine-tuning specialized, open-source VLMs for time series visualization evaluation represents a crucial future direction.
Additionally, the proposed on \texttt{TimeVista} focuses on univariate evaluation, rendering each prediction as a single 2D line plot. Extending the VLM-as-a-Judge paradigm to multivariate scenarios, such as multi-sensor monitoring or multi-asset portfolios, poses non-trivial challenges in visualization design and cross-variable reasoning. We leave a systematic investigation of multivariate evaluation as an important direction for future work.
Lastly, while our macro-level evaluation successfully incorporates rich domain knowledge, general-purpose VLMs may still struggle with highly specialized, private, or rapidly evolving expert contexts, such as rare industrial anomalies or sudden financial black-swan events. Fully capturing these dynamic and highly niche scenarios remains an open challenge that requires the continuous expansion of specialized, expert-curated rubrics.

\bibliography{custom}

\appendix
\section{Ethical Considerations}
We state that all existing datasets and APIs used in this study were employed strictly for academic research and evaluation, fully consistent with their intended use and terms of service, with the proposed benchmark's intended use detailed in Section 5. 
Our human annotation protocol, which was determined to be exempt from formal ethics review by our Institutional Review Board (IRB) due to the absence of sensitive personal data or physical/psychological risks, involved active time series researchers recruited from universities. 
All participants provided formal, written consent after being explicitly informed of the study's purpose and data usage, and they were compensated adequately and fairly relative to local standards; the full text of the annotation instructions is provided in Appendix~\ref{app:rubrics}. 
Finally, we declare that AI assistants were used solely for polishing the language and improving the readability of the manuscript, while all core scientific contributions, experimental designs, and writing were completed by the human authors.

\section{Data Curation}

We specify that all datasets used in this work are open-source and utilized in strict accordance with their respective official licenses. To ensure ethical and legal compliance, we have carefully verified that all data sources are publicly available for research purposes, contain no personally identifiable information (PII), and respect intellectual property rights. The curation process was conducted solely to facilitate fair and reproducible benchmarking in time series forecasting.

\begin{table*}[htbp]
\caption{Individual statistics of \texttt{TimeVista} across all datasets.}
\centering
\small 
\begin{tabular}{llcclc}
\toprule
\textbf{Dataset} & \textbf{Domain} & \textbf{Input Length} & \textbf{Prediction Length} & \textbf{Frequency} & \textbf{\# Series}  \\

\midrule

CPSC2019 & Medical & 500 & 100 & S & 299 \\

PTB & Medical & 500 & 100 & S & 300 \\

RDBH & Medical & 500 & 100 & T & 300 \\

FNSPID & Finance & 42 & 7 & D & 500 \\

CPHL & Nature & 1818 - 17542 & 12, 336 & 15T, 30T, H & 494 \\

Supply Chain Customer & Sales & 1642 - 1972 & 30 & D & 424 \\

Australia Solar & Energy & 32544 - 35040 & 24, 168 & H & 448 \\

Azure2019 & Web/CloudOps & 594 - 8351 & 48, 288 & 5T & 500 \\

Crypto & Finance & 2569 - 2809 & 30 & D & 29 \\

Current Velocity & Nature & 1536 - 31849 & 9, 432 & 10T, 15T, 20T, 5T, H & 500 \\

Covid Death & HealthCare & 200 - 468 & 21 & D & 569 \\

Loop Seattle & Transportation & 7944 - 104256 & 48, 480 & 5T, H & 400 \\

Solar & Energy & 8328 - 8712 & 48 & H & 400 \\

Synthetic & Synthetic & 107 - 1720 & 32, 601 & - & 400 \\

\bottomrule

\end{tabular}

\label{tab:dataset_stats}

\end{table*}


\subsection{Real-world Dataset}

\paragraph{Azure2019~\cite{qiao2026s}} Sourced from the Azure cloud platform, this dataset contains processed CPU utilization traces at 5-minute intervals. The data are categorized into three workload classes—Delay-insensitive (D), Interactive (I), and Unknown (U)—where categories D and I exhibit significantly longer temporal durations than category U.

\paragraph{Covid Death~\cite{dong2020interactive}} 
The dataset is sourced from the open-source repository maintained by Johns Hopkins University. It compiles global epidemiological data on the 2019 novel coronavirus (COVID-19), a pathogen identified as the causative agent of a widespread respiratory illness outbreak. The dataset provides daily-level time-series records of deaths spanning from January 22, 2020, to May 29, 2021, with all metrics reported as cumulative figures up to each specific observation date. The geographical coverage encompasses various countries and regions worldwide, with finer-grained spatial resolution down to the province, state, or territory level for certain areas. This comprehensive dataset provides robust empirical support for analyzing the spatiotemporal transmission dynamics and evolutionary patterns of the pandemic.

\paragraph{CPHL~\cite{qiao2026s}} This dataset monitors chlorophyll mass concentration across four marine stations in Australia, with raw sampling frequencies spanning 15- to 30-minute intervals. These records were downsampled to a standardized hourly resolution to ensure temporal alignment.

\paragraph{Crypto~\cite{qiao2026s}} This dataset comprises historical daily price records for four major cryptocurrencies—Bitcoin, Ethereum, Litecoin, and Bitcoin Cash—spanning the period from December 20, 2017, to September 30, 2025.

\paragraph{ECG~\cite{tang2025ecg}}
The electrocardiogram (ECG) data originates from three sources: the CPSC2019 dataset, which contains 2,000 single-lead ECG recordings collected from patients diagnosed with cardiovascular disease; the PTB and PTB-XL Diagnostic ECG Databases, where PTB contains 549 records from 290 subjects, and the larger PTB-XL cohort provides 21,799 clinical 12-lead ECGs (each lasting 10 seconds) from 18,869 patients; and the Detection of Heart Beats Dataset (RDBH), which consists of longer-duration, 10-minute recordings containing four to eight signals per record. Together, these datasets offer a comprehensive spectrum of lead configurations, recording durations, and clinical conditions, establishing a robust foundation for evaluating ECG forecasting capability.

\paragraph{FNSPID~\cite{dong2024fnspid}}
This dataset originates from The Financial News and Stock Price Integration Dataset (FNSPID), a comprehensive, large-scale financial benchmark designed to enhance stock market predictions by integrating quantitative market data with qualitative textual information. Spanning from 1999 to 2023, FNSPID contains 29.7 million stock price records and 15.7 million financial news articles covering 4,775 S\&P 500 companies.

\paragraph{Loop Seattle~\cite{cui2018loop_seattle}}
This dataset is a station-based traffic dataset, referred to as the loop detector data, which was collected via inductive loop detectors embedded in the roadway surface. In this infrastructure, multiple loop detectors are connected to a single detector station, with these stations deployed approximately every half-mile. The raw data collected from individual detectors are grouped and aggregated by directional flow to represent station-based traffic states; for the purposes of this study, only the speed measurements were retained. Geographically, the dataset covers four interconnected freeways in the Seattle metropolitan area—namely, I-5, I-405, I-90, and SR-520. These data were extracted from the Digital Roadway Interactive Visualization and Evaluation Network (DRIVE Net) system.

\paragraph{Solar~\cite{lai2018solar}}
This dataset comprises solar power production records from the year 2006, which were sampled at 10-minute intervals from 137 photovoltaic (PV) plants located across the State of Alabama. This high-resolution dataset provides a continuous and detailed representation of solar energy generation, serving as a reliable basis for evaluating the temporal modeling capabilities of forecasting methods.

\begin{table*}[t]
\centering
\caption{Model configurations of TSFMs used in evaluation. Context length refers to the maximum number of historical time steps used as input. Input mode indicates whether the model natively supports multimodal input. Output type specifies whether models produce quantile forecasts or distribution-based probabilistic forecasts.}
\label{tab:model_config}
\resizebox{\textwidth}{!}{%
\begin{tabular}{llcllc}
\toprule
\textbf{Model} & \textbf{Checkpoint} & \textbf{Context Length} & \textbf{Input Mode} & \textbf{Output Type} \\
\midrule
Timer-S1 & \href{https://huggingface.co/bytedance-research/Timer-S1}{\texttt{bytedance-research/Timer-S1}} & 11520 & Unimodal & Quantile  \\
TimesFM 2.5 & \href{https://huggingface.co/google/timesfm-2.5-200m-pytorch}{\texttt{google/timesfm-2.5-200m-pytorch}} & 4096 & Unimodal & Quantile  \\
Chronos-2 & \href{https://huggingface.co/amazon/chronos-2}{\texttt{amazon/chronos-2}} & 8192 & Unimodal & Quantile  \\
Moirai 2.0 & \href{https://huggingface.co/Salesforce/moirai-2.0-R-small}{\texttt{Salesforce/moirai-2.0-R-base}} & 4000 & Unimodal & Quantile  \\
TiRex & \href{https://huggingface.co/NX-AI/TiRex}{\texttt{NX-AI/TiRex}} & 2048 & Unimodal & Quantile  \\
Sundial & \href{https://huggingface.co/thuml/sundial-base-128m}{\texttt{thuml/sundial-base-128m}} & 2880 & Unimodal & Distribution \\
Chronos & \href{https://huggingface.co/amazon/chronos-bolt-base}{\texttt{amazon/chronos-bolt-base}} & 2048 & Unimodal & Quantile \\
FlowState & \href{https://huggingface.co/ibm-research/flowstate}{\texttt{ibm-research/flowstate}} & 4096 & Unimodal & Quantile \\
Aurora & \href{https://huggingface.co/DecisionIntelligence/Aurora}{\texttt{DecisionIntelligence/Aurora}} & 2048 & Multimodal & Distribution \\
\bottomrule
\end{tabular}%
}
\end{table*}

\paragraph{Australia Solar~\cite{qiao2026s}} This dataset consists of hourly-resolution solar power generation records collected from three distinct photovoltaic (PV) installations across Australia.

\paragraph{Current Velocity~\cite{qiao2026s}} This dataset comprises time-series observations of ocean current velocities from the National Mooring Network Facility, which monitors coastal waters across five Australian states. The dataset includes primary flow indicators—zonal (UCUR), meridional (VCUR), and vertical (WCUR) velocity components—alongside key hydrographic parameters: seawater temperature (TEMP), pressure (PRES), and sound speed (SSPD). To harmonize the varying minute-level sampling frequencies across different stations, the raw records were downsampled to an hourly resolution, with excessively short sequences excluded to ensure data continuity.

\paragraph{Supply Chain Customer~\cite{qiao2026s}} This dataset contains five years of simulated transactional delivery records. Daily metrics, including order counts, piece volumes, and total revenue, are aggregated at the customer level.

\begin{table*}[t]
\centering
\caption{Inter-Annotator Agreement (IAA) across different evaluation dimensions.}
\label{tab:IAA_result}
\setlength{\tabcolsep}{9pt} 
\begin{tabular}{lcccccc}
\toprule
\multirow{2}{*}{\textbf{Metric}} & \multicolumn{4}{c}{\textbf{Micro-level}} & \multicolumn{2}{c}{\textbf{Macro-level}}\\
\cmidrule(lr){2-5} \cmidrule(lr){6-7}
& Trend & Motif & Phase & Magnitude & w/o knowledge & w/ knowledge \\
\midrule
Krippendorff's $\alpha$ & 0.679 & 0.665 & 0.677 & 0.728 & 0.654 & 0.891 \\
Kendall's $\tau$        & 0.889 & 0.876 & 0.899 & 0.906 & 0.898 & 0.942 \\
Spearman's $\rho$       & 0.907 & 0.895 & 0.917 & 0.920 & 0.916 & 0.961 \\
\bottomrule
\end{tabular}
\end{table*}

\subsection{Synthetic Data}
\label{app:synthesis}
To isolate and comprehensively evaluate the VLM's structural perception, we instantiate Eq.~\ref{time_series_decomposition} by using the \textit{kernel-synth} method \cite{ansari2025chronos}. Treating the four morphological dimensions as independent parameters, this orthogonal design enables targeted interventions without unpredictable feature coupling, yielding a noise-free ground truth. To prevent the synthesis from collapsing into overly simplistic patterns, we integrate Gaussian Processes (GP) with composite kernels for capturing complex macro trends, alongside time-varying modulators for stochastic amplitude fluctuations. This synthesizes a diverse spectrum of complex temporal dynamics, balancing rigorous control with realistic comprehensiveness. Beyond evaluating the individual components, we introduce a \textit{General} dimension to assess the model's overall structural fidelity.

\section{Implementation Details}

\subsection{Models}

We evaluate 13 time series forecasting models, including 9 time series foundation models, using their official checkpoints from HuggingFace. Table~\ref{tab:model_config} summarizes the key hyperparameters used in our evaluation. All models are evaluated in a zero-shot setting without any fine-tuning.

\subsection{Metrics}

\paragraph{Mean Squared Error (MSE).} MSE measures the average of the squared differences between the predicted values and the ground truth across the forecasting horizon. By squaring the errors, MSE heavily penalizes larger deviations, making it highly sensitive to outliers and large forecast errors. It is defined as:
\begin{align}
\text{MSE} &= \frac{1}{H} \sum_{i=1}^H \left( y_i - \widehat{y}_i \right)^2,
\end{align}
where $y_i$ and $\widehat{y}_i$ denote the ground truth and predicted value at the $i$-th future time step, respectively.

\paragraph{Mean Absolute Error (MAE).} MAE computes the average of the absolute differences between the ground truth and the predicted values. Unlike MSE, MAE provides a linear penalty for errors, offering a more robust and direct reflection of the average forecast error magnitude without being overly sensitive to outliers. It is defined as:
\begin{align}
\text{MAE} &= \frac{1}{H} \sum_{i=1}^H \left| y_i - \widehat{y}_i \right|,
\end{align}
where $y_i$ represents the target value at the $i$-th future time step.

\paragraph{Dynamic Time Warping (DTW).} DTW assesses temporal similarity by finding an optimal alignment path between the ground truth and predicted sequences, accommodating potential temporal shifts or phase distortions. This metric is crucial for evaluating whether a model successfully captures the underlying structural shape and profile of the time series rather than relying solely on rigid point-wise alignment. It is defined via the optimization problem:
\begin{align}
\text{DTW}(\mathbf{y}, \widehat{\mathbf{y}}) &= \min_{W} \sum_{k=1}^K \delta(w_k),
\end{align}
where $W = \langle w_1, w_2, \dots, w_K \rangle$ represents a warping path of length $K$. Each element $w_k = (i_k, j_k)$ signifies an alignment between the $i_k$-th step of $\mathbf{y}$ and the $j_k$-th step of $\widehat{\mathbf{y}}$. For univariate series, the distance function $\delta(w_k)$ computes the squared absolute difference between the scalar values, defined as $\delta(w_k) = (y_{i_k} - \widehat{y}_{j_k})^2$.

\paragraph{Mean Absolute Scaled Error (MASE).} MASE provides a scale-independent assessment of forecast accuracy by normalizing the prediction error against the mean absolute error of a seasonal naive baseline. This normalization ensures interpretability across different time series and granularities. It is defined as:
\begin{align}
\text{MASE} &= \frac{1}{H} \sum_{i=1}^H \frac{|y_i - \widehat{y}_i|}{\frac{1}{H-s} \sum_{j=s+1}^{H} |y_j - y_{j-s}|},
\end{align}
where $s$ denotes the periodicity of the data (e.g., season length), and $H$ represents the forecasting horizon. $\mathbf{y}, \widehat{\mathbf{y}} \in \mathbb{R}^{H}$ denote the ground truth and the predicted value sequences, respectively, and $y_i$ refers to the value at the $i$-th future time step.

\section{More Experimental Results}

\subsection{IAA}

To ensure ethical data collection, all annotators were compensated with a fair wage that strictly aligns with the local standards for intellectual annotation tasks. Prior to starting the annotation task, all participants were explicitly informed about the purpose of our study, the nature of the tasks, and how their annotated data would be used. They provided formal consent prior to their participation, acknowledging their voluntary involvement and agreeing to the academic use of their contributions.

To validate the quality and consistency of the human annotations, we calculate inter-annotator agreement (IAA) using multiple robust metrics: Krippendorff's $\alpha$, Kendall's $\tau$, and Spearman's $\rho$. As detailed in Table \ref{tab:IAA_result}, the results reveal a consensus among annotators across both micro-level and macro-level settings. These high agreement scores indicate that our evaluation task is well-defined, the rubrics are unambiguous, and the resulting human ground truth is highly reliable for benchmarking the VLM judges.


\subsection{Consistency}

\begin{table}[htbp]

\centering

\small

\caption{Consistency analysis of the VLM-as-judge framework across 5 independent runs.}

\label{tab:vlm_consistency}

\begin{tabular}{ccc}

\toprule

\textbf{Metric / Dimension} & \textbf{Count / Value} & \textbf{Percentage (\%)} \\

\midrule

\multicolumn{3}{l}{\textbf{Overall Statistics}} \\

Total Samples & 2,945 & 100.00 \\

Exact Match & 2,141 & 72.70 \\

Average std & 0.1594 & -- \\

\midrule

\multicolumn{3}{l}{\textbf{Score Difference}} \\

Difference = 0 & 2,141 & 72.70 \\

Difference = 1 & 663 & 22.51 \\

Difference = 2 & 112 & 3.80 \\

Difference = 3 & 23 & 0.78 \\

Difference = 4 & 6 & 0.20 \\

\bottomrule

\end{tabular}

\end{table}

To assess reproducibility, we analyze the scoring stability of the VLM judge across five independent runs. The framework exhibits remarkable consistency, achieving an overall exact match rate of 72.70\% ($\sigma = 0.0$) and a low average standard deviation of 0.159. Crucially, 95.21\% of the samples show a maximum score discrepancy of at most 1.0 (with 22.51\% having a 1-point variance), while severe fluctuations ($\ge 3.0$) occur in less than 1\% of cases. These results highlight the robust reliability and reproducibility of VLM-based assessments.

\subsection{Sensitivity}

We evaluate the VLM judge's robustness to geometric scaling by varying the plot aspect ratio ($x$:$y$) across four configurations: vertically compressed (\texttt{6:5}), default (\texttt{12:5}), and horizontally elongated (\texttt{18:5} and \texttt{24:5}). 
As shown in Table~\ref{tab:aspect_ratio_sensitivity}, the VLM judge demonstrates remarkable invariance to these geometric changes. Both Spearman's $\rho$ and Kendall's $\tau$ remain exceptionally stable across all settings, fluctuating within a marginal range of merely $0.011$ and $0.015$, respectively. Even under extreme horizontal stretching (\texttt{24:5}) or vertical compression (\texttt{6:5}), the VLM consistently maintains strong alignment with human preferences. This confirms that the VLM's visual evaluation is driven by intrinsic geometric relationships and temporal trends rather than absolute rendering dimensions.
\begin{table}[h]
\centering
\setlength{\tabcolsep}{12pt} 
\renewcommand{\arraystretch}{1.2} 
\small 

\caption{Sensitivity analysis of VLM judges under different plot aspect ratios ($x$:$y$ axis ratio).}
\vspace{-5pt}

\begin{tabular}{l|cc}
\toprule
\textbf{Aspect Ratio} & \textbf{$\rho$} & \textbf{$\tau$} \\
\toprule
6:5  & 0.896 & 0.836 \\
12:5 (Default) & 0.907 & 0.849 \\
18:5 & 0.903 & 0.846 \\
24:5 & 0.907 & 0.851 \\
\bottomrule
\end{tabular}
\label{tab:aspect_ratio_sensitivity}
\vspace{-5pt}
\end{table}

\subsection{Contextual Re-interpretation: Ablation of Knowledge}

A fundamental advantage of VLMs is their potential to dynamically adjust evaluative focus on identical visual inputs by grounding them in textual domain constraints. To verify this, we compare the VLM's judging behavior on the real-world subset of \texttt{Meta-TimeVista} across the two macro-level settings and analyze the impact of domain knowledge on score variation.

\begin{figure}[h]
  \centering
  \includegraphics[width=\columnwidth]{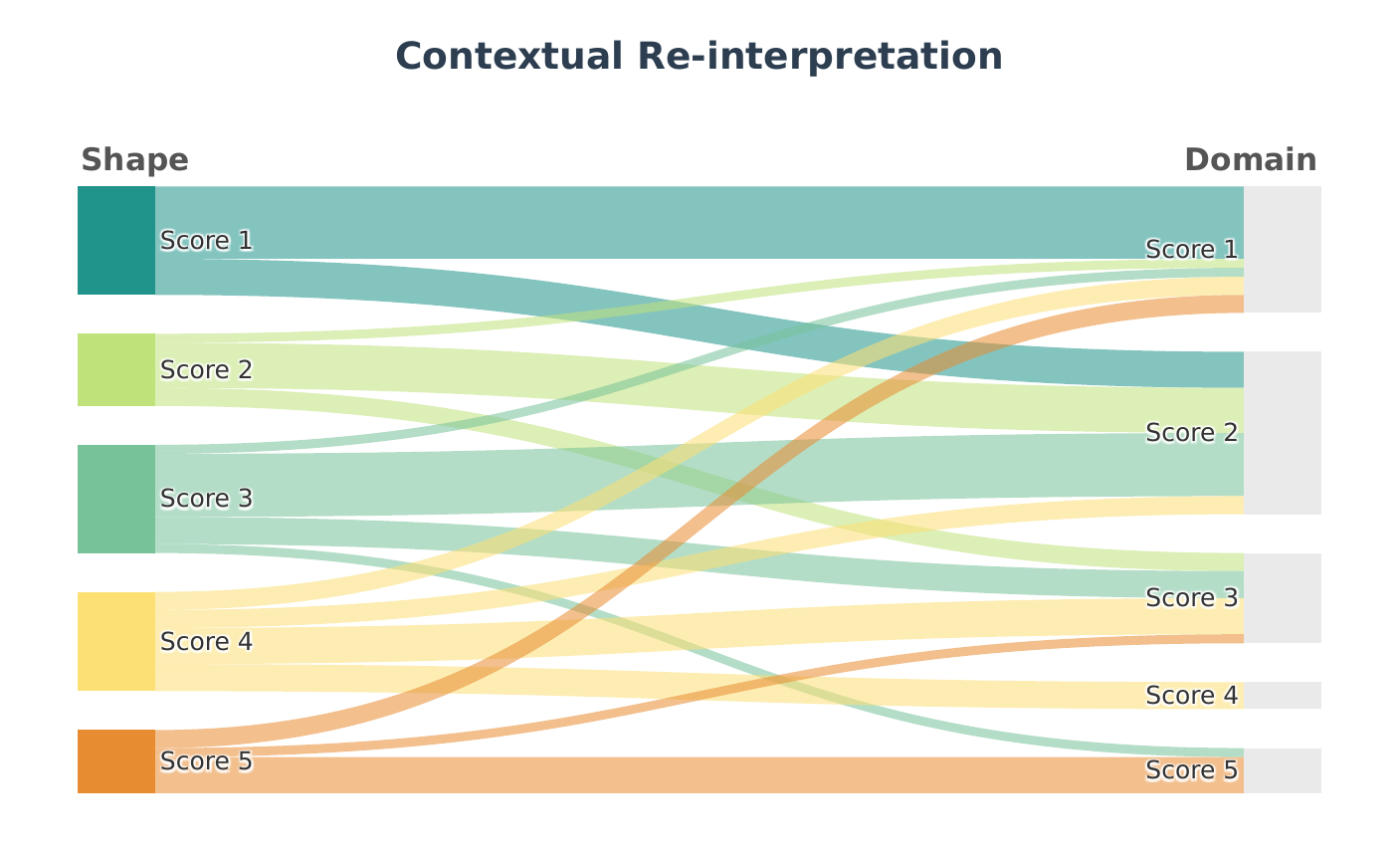}
  \caption{Sankey diagram illustrating the dynamic flow of VLM scores when transitioning from Macro (w/o knowledge) to context-aware Macro (w/ knowledge).}
  \label{fig:reinterpretation}
\end{figure}

Importantly, Macro (w/o knowledge) (assessing geometric fidelity) and Macro (w/ knowledge) (assessing practical utility) are not inherently superior to one another; they simply represent different valid perspectives depending on downstream needs. The true power of the VLM lies in its capacity to seamlessly switch between them. As visualized in Figure \ref{fig:reinterpretation}, introducing domain context triggers a profound re-distribution of scores. We observe complex crossing flows: forecasts penalized for minor phase shifts under Macro (w/o knowledge) are frequently upgraded if the domain tolerates such deviations, whereas visually smooth predictions are downgraded if they violate critical business constraints. This fluid score modulation demonstrates that our VLM-as-a-Judge is not a rigid formula, but a highly flexible, multi-dimensional judge capable of adapting to diverse real-world scenarios.

\subsection{Analysis of VLM Scoring Preferences}

\begin{figure}[h]
  \centering
  \includegraphics[width=\columnwidth]{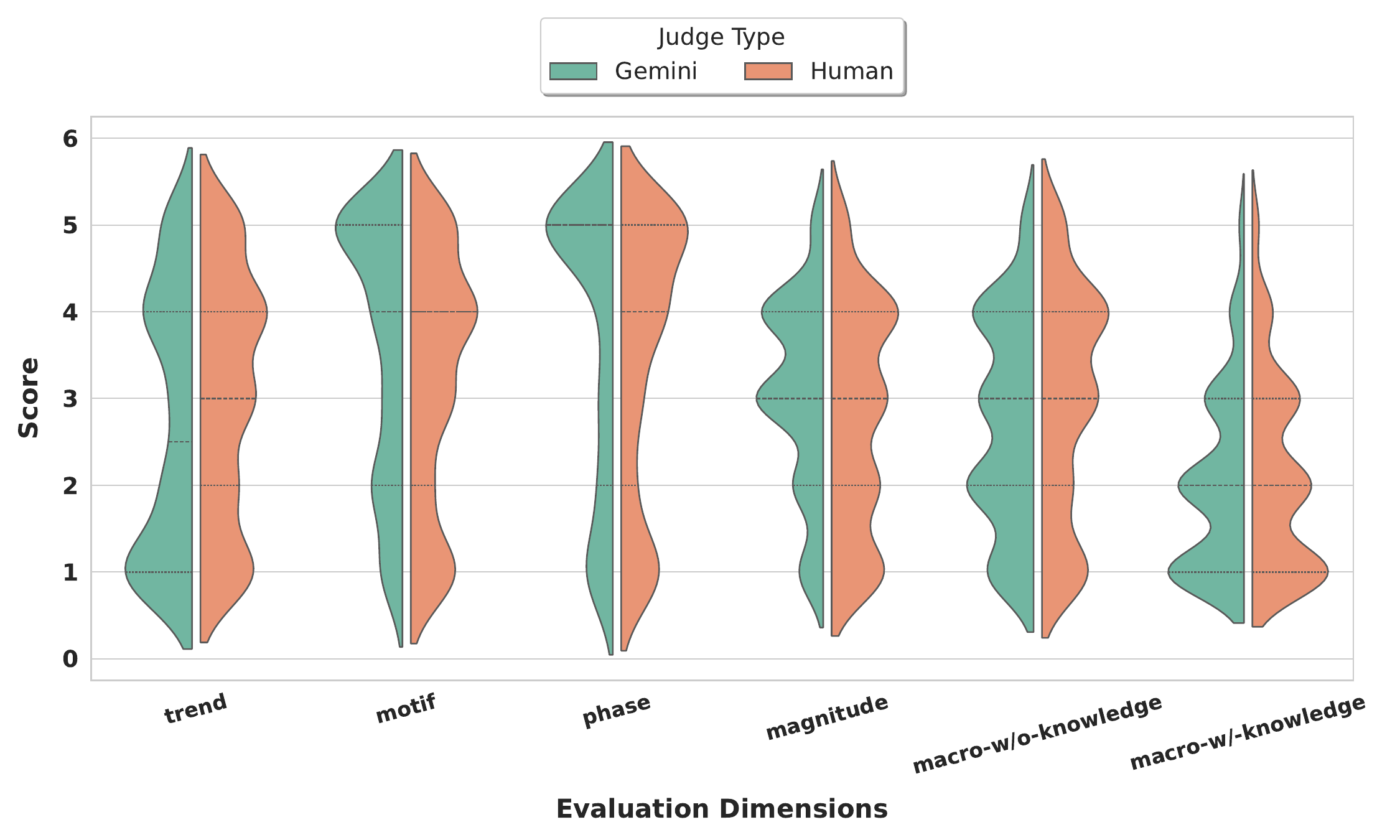}
  \vspace{-10pt}
  \caption{Comparison of score distributions between Gemini-3.1-Pro and human annotators across micro- and macro-level evaluation dimensions.}
  \vspace{-20pt}
  \label{fig:vlm_score_distribution}
\end{figure}

To understand the decision-making behavior of the VLM judge, we analyze the score distributions assigned by Gemini-3.1-Pro and human annotators across different evaluation dimensions, as illustrated in Figure~\ref{fig:vlm_score_distribution}. Overall, Gemini's scoring distributions align remarkably well with human annotations across almost all dimensions, demonstrating that the VLM judge captures evaluation criteria that closely mimic human intuition. 

In the micro-level dimensions, both judges exhibit highly similar patterns. For instance, on \textit{motif} and \textit{phase}, both Gemini and humans show a clear top-heavy distribution with medians clustered around 4 and 5, indicating a shared tolerance or standard for local pattern matching. On \textit{magnitude}, both distributions display a similar multi-modal shape, peaking at scores of 1, 3, and 4. The most compelling alignment is observed in the macro-level evaluations. In the \textit{macro-w/o-knowledge} setting, both Gemini and humans present a balanced, multi-modal distribution centered around a median of 3. However, once domain-specific constraints are introduced (\textit{macro-w/-knowledge}), both distributions shift dramatically downward, with the medians dropping to 2 and a high concentration of scores pooling at the lower end. This shared shift demonstrates that, like humans, the VLM judge becomes significantly more stringent when evaluating physical and operational feasibility, proving its capability to enforce domain-specific constraints rather than relying solely on visual heuristics.

\section{Evaluation Cases}

\subsection{VLM Scoring and Rationale: Case Studies}\label{sec:app_example}
In the main text, we provide an example of Gemini-3.1-Pro judging performance. In this section, we provide more examples.

\begin{figure*}[t]
  \centering
  \includegraphics[width=0.85\linewidth]{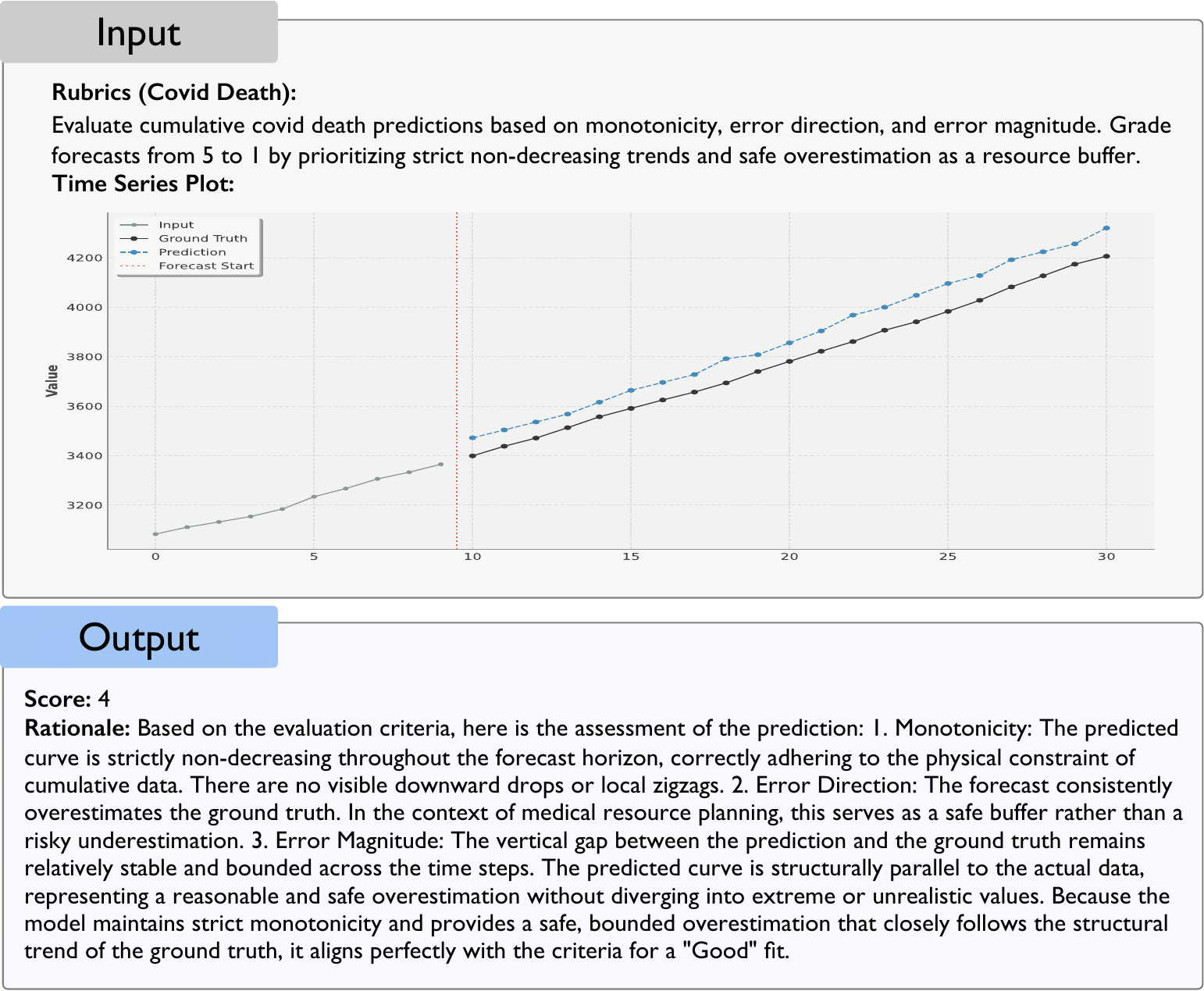}
  \label{fig:case_covid}
  \caption{An example of judgment from Gemini-3.1-Pro for covid death prediction.}
\end{figure*}

\begin{figure*}[t]
  \centering
  \includegraphics[width=0.85\linewidth]{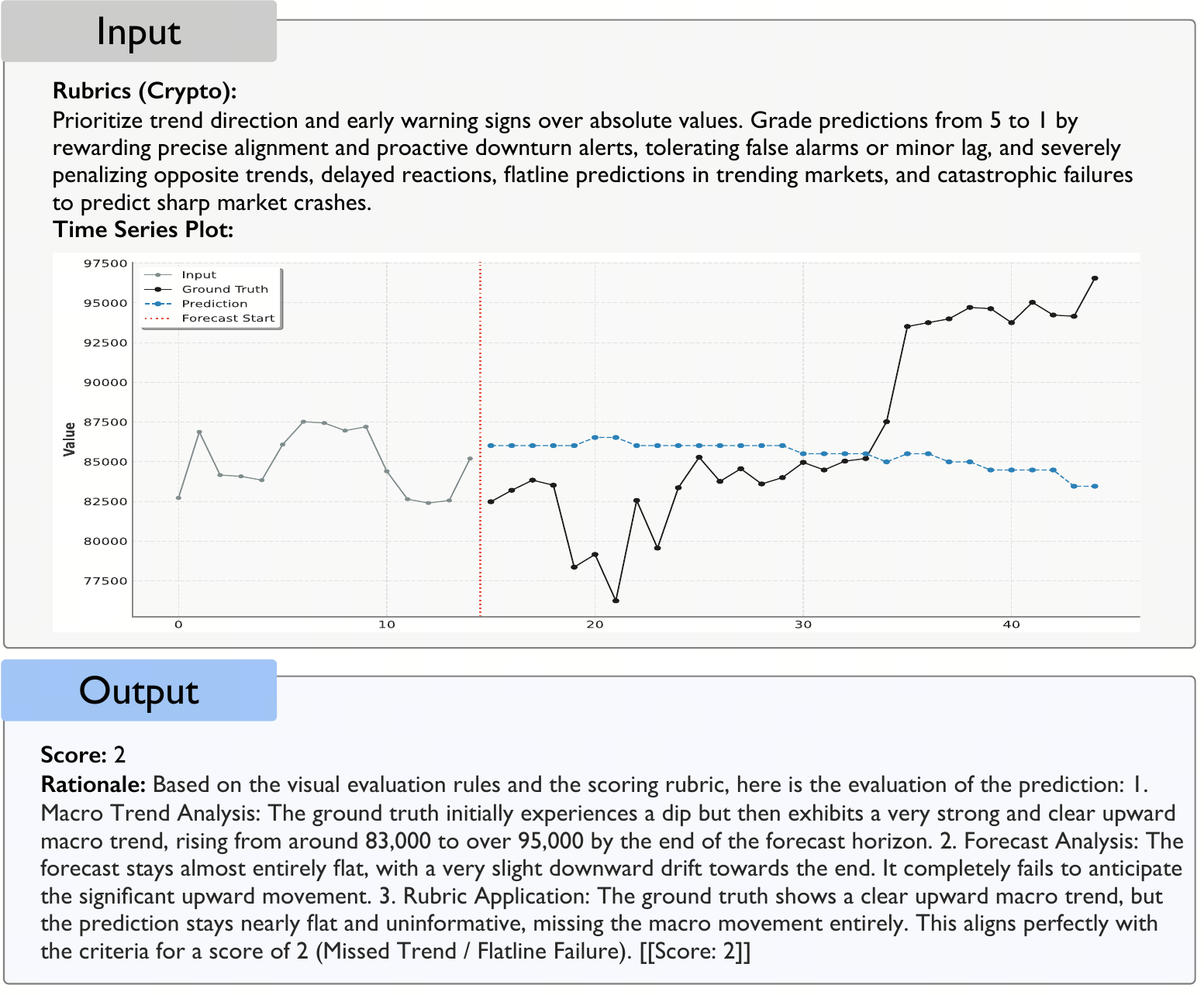}
  \label{fig:case_crypto}
  \caption{An example of judgment from Gemini-3.1-Pro for crypto prediction.}
\end{figure*}


\begin{figure*}[t]
  \centering
  \includegraphics[width=0.85\linewidth]{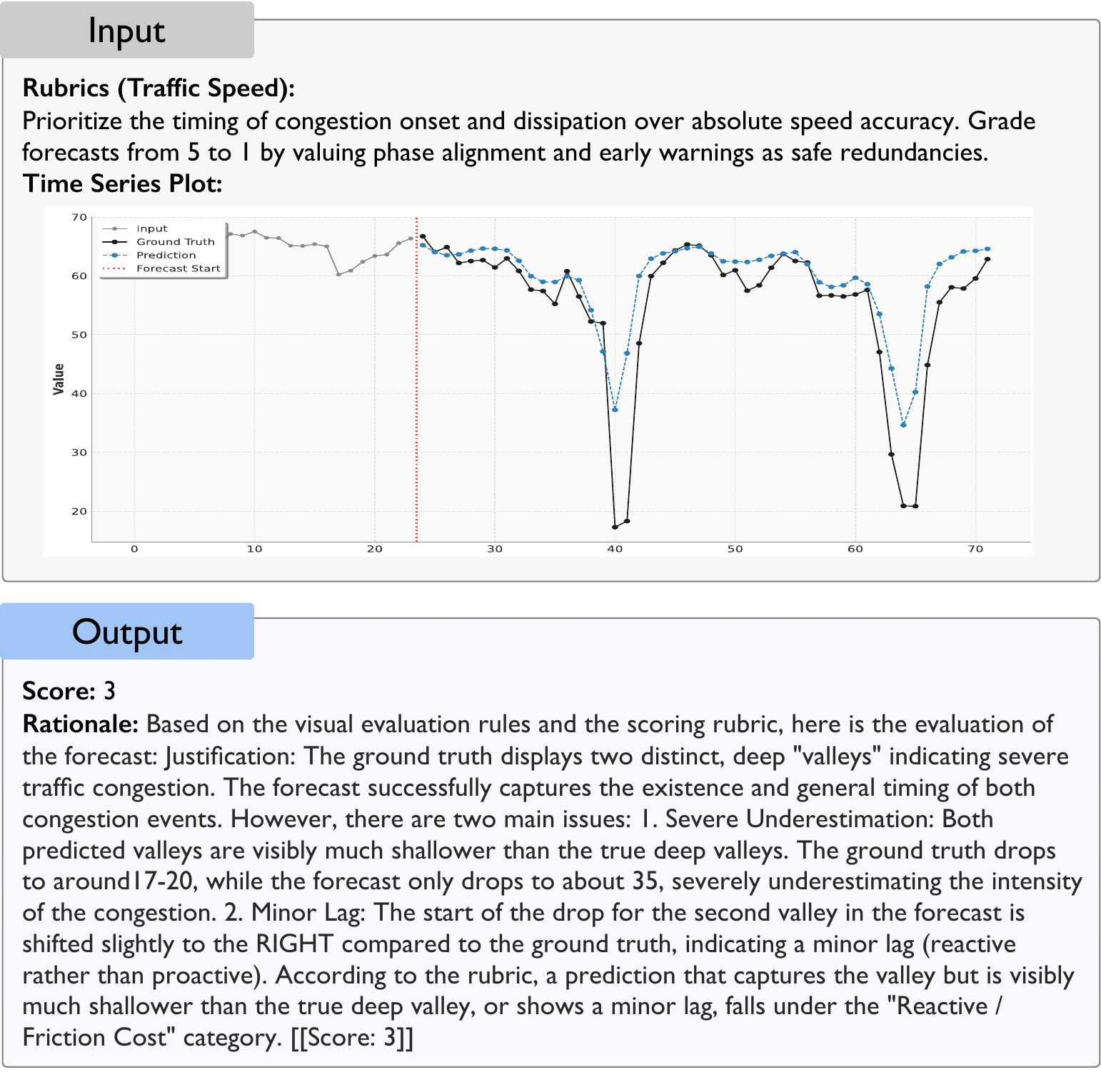}
  \label{fig:case_speed_1}
  \caption{An example of judgment from Gemini-3.1-Pro for traffic speed prediction.}
\end{figure*}

\begin{figure*}[t]
  \centering
  \includegraphics[width=0.85\linewidth]{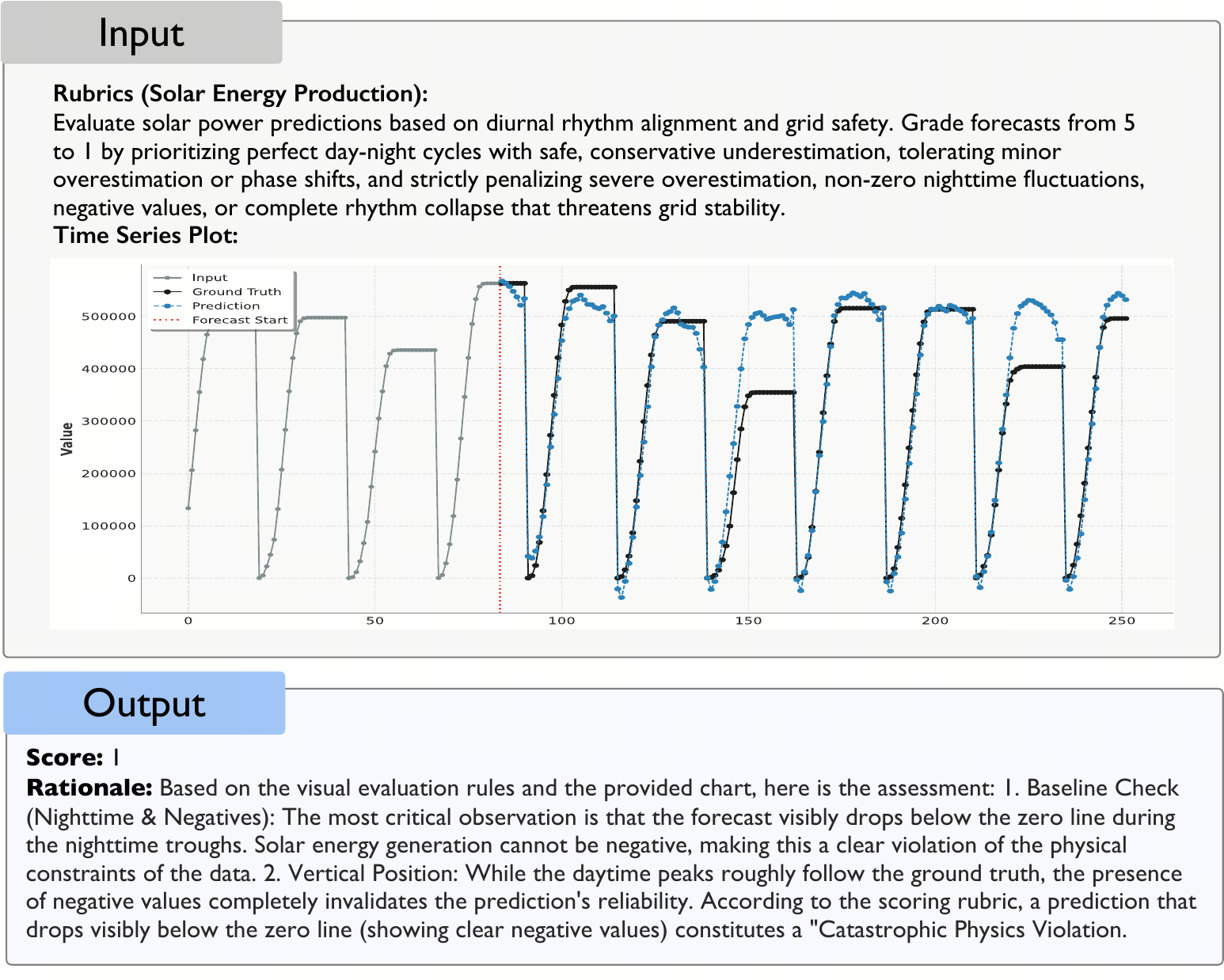}
  \label{fig:case_solar}
  \caption{An example of judgment from Gemini-3.1-Pro for solar energy production prediction.}
\end{figure*}

\subsection{Advantage over Traditional Metrics: Case Studies}
To illustrate why traditional point-wise metrics often fail to reflect practical utility, we present six representative case studies from our macro-level evaluation (w/o knowledge). These examples highlight the discrepancy between numerical errors and human judgments.
In these cases, traditional metrics such as MSE or MAE frequently identify a model as the top performer simply because its predictions have a smaller average distance to the ground truth. However, such numerical alignment does not always translate to a useful forecast. For instance, a model can achieve a low MSE by predicting a flat, near-constant line that entirely misses the temporal dynamics of the sequence. Conversely, another model that successfully captures the overall trend and peak-valley motifs might be heavily penalized by point-wise metrics due to a minor constant offset or a slight phase shift.
Furthermore, traditional metrics are incapable of detecting qualitative errors, such as predicting negative values for strictly positive physical quantities. While MSE remains indifferent to these unrealistic predictions, both human experts and the VLM judge identify and penalize them. 
Ultimately, these case studies demonstrate that evaluating time series forecasting requires more than calculating point-to-point distances. The VLM judge aligns closely with human experts by evaluating the structural integrity of the predictions, tolerating benign numerical deviations, and focusing on whether the predicted patterns are physically and operationally plausible.

\begin{figure*}[t]
    \centering
  \includegraphics[width=0.9\linewidth
  ]{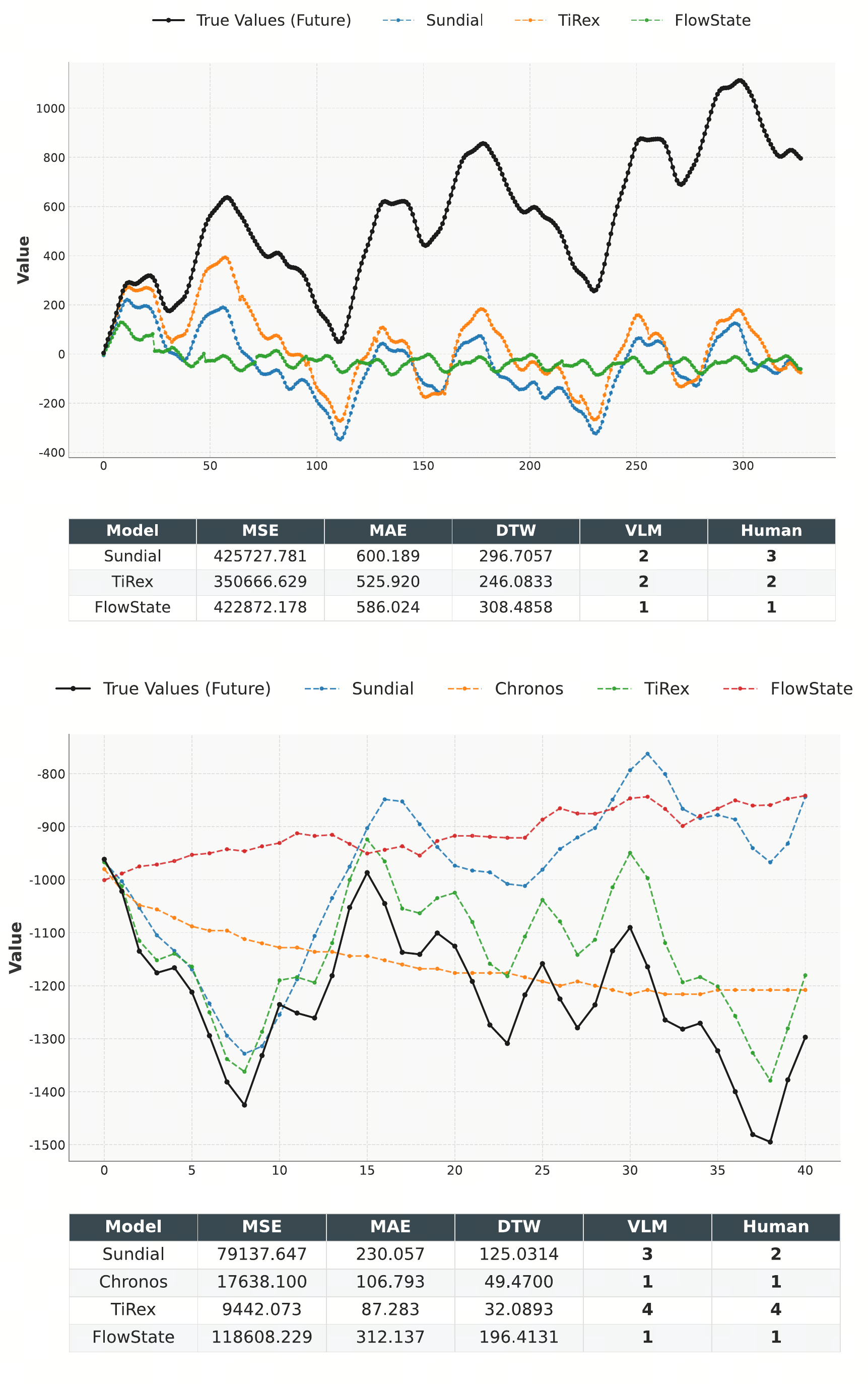}
  \label{fig:case1}
  \vspace{-10pt}
\end{figure*}

\begin{figure*}[t]
    \centering
  \includegraphics[width=0.9\linewidth
  ]{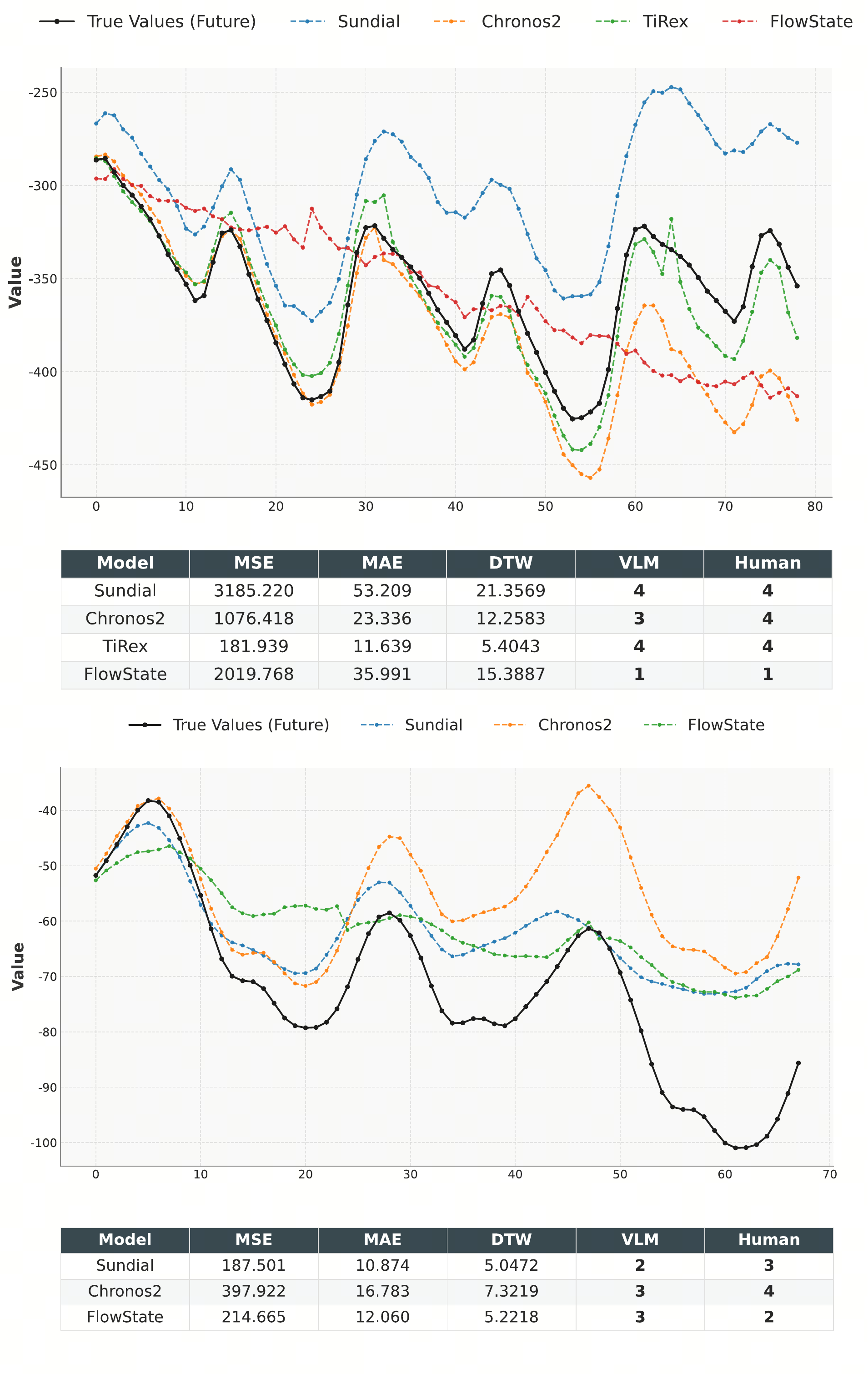}
  \label{fig:case2}
  \vspace{-10pt}
\end{figure*}

\begin{figure*}[t]
    \centering
  \includegraphics[width=0.9\linewidth
  ]{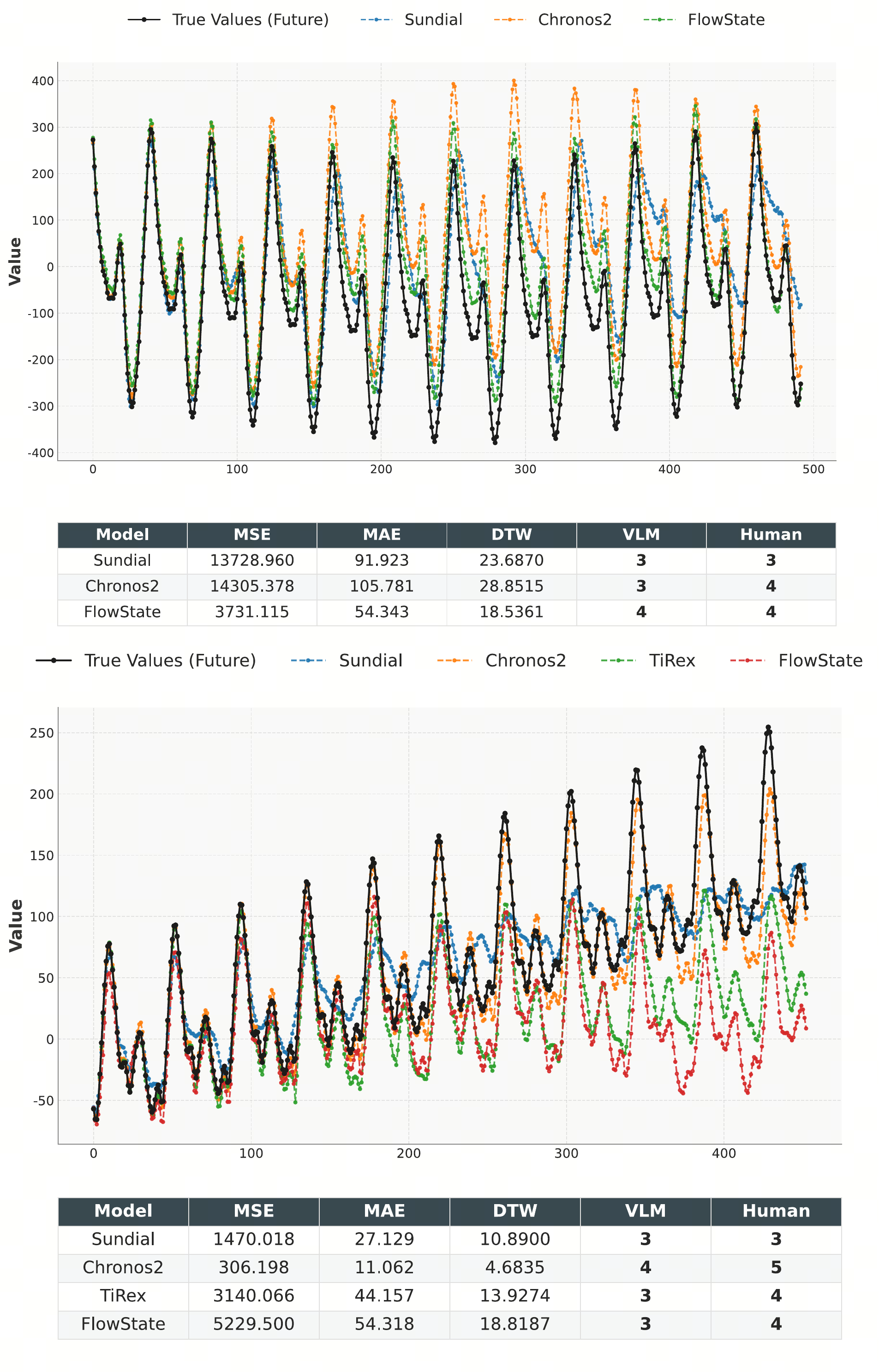}
  \label{fig:case3}
  \vspace{-10pt}
\end{figure*}

\section{Prompt}

\label{app:rubrics}
\subsection{System Prompt}

\begin{tcolorbox}[float=h]
    \textbf{System Prompt}
    \tcblower 
    You are an expert evaluator in time series forecasting. You will be provided with a chart displaying historical data, ground truth, and model predictions. Your task is to evaluate the prediction quality based on a specific dimension and assign a score from 1 to 5. Please output your reasoning and the score strictly in the format:    
    [[Score: X]], where X is an integer from 1 to 5.
\end{tcolorbox}

\subsection{User Prompt}

\subsubsection{Micro-level Evaluation}

\begin{tcolorbox}[float=h]
    \textbf{User Prompt: Trend}
    \tcblower 
    \small
    Evaluate the consistency of the [Macro Trend] (the overall baseline trajectory and major turning points). \\
    \textbf{Visual Evaluation Rubrics:} \\
    Imagine a smoothed ``center of gravity'' line for both curves. Compare these macro paths.You may IGNORE minor local zig-zag noise or recurring periodic waves \textbf{IF AND ONLY IF} they are small in amplitude and ride on top of a clear, discernible macro-trend line. \textbf{HOWEVER}, if the noise itself is of high amplitude and dominates the visual appearance of the prediction, obscuring the underlying trend, then you \textbf{MUST NOT} ignore it; instead, it should be heavily penalized according to the rubric's descriptions for scores 1-2. FOCUS ONLY on the overall direction (up/down/flat) and the timing of major macro bends/turns. \\

    \textbf{Scoring Rubric:} \\
    \textbf{5 (Perfect Fit):} The overall visual baselines, movement directions, and major turning points align perfectly. \\
    \textbf{4 (Structurally Intact with Minor Flaws):} Captures the primary direction and major turns. For monotonic trends, the slope angle has a minor deviation (the vertical gap between the two curves at the end of the chart is SMALLER than the total vertical height the ground truth climbed/dropped). For complex trends, a major turn happens slightly earlier or later. \\
    \textbf{3 (Noticeable Trend Degradation):} Matches the trend in some parts but has severe visual flaws. Severe slope distortion (the vertical gap at the end is LARGER than the ground truth's total height change), or it visually misses a secondary macro bend/turn.  \\
    \textbf{2 (Significant Deviation or Excessive Noise):} For 30\%-50\% of the horizontal axis, the prediction moves in the completely opposite direction (e.g., goes up while GT goes down) or flattens out horizontally (mean reversion). OR, hallucinates a mild slope when GT is almost visually flat. \textbf{OR, the prediction is visually dominated by high-frequency, seemingly random noise, making its underlying macro trend difficult to discern or unreliable, even if some directional changes are occasionally matched.} \\
    \textbf{1 (Completely Contradictory or Degenerated to Pure Noise):} For $>$50\% of the horizontal axis, the curves move in opposite directions (e.g., forming an ``X'' shape) or the prediction globally flattens out. OR, hallucinates a strong, massive trend on a completely flat/stable ground truth. \textbf{OR, the prediction has completely degenerated into pure, seemingly random noise, making any underlying trend entirely unidentifiable or meaningless across the majority of the forecast horizon. It offers no reliable macro directional information.}
\end{tcolorbox}

\begin{tcolorbox}[float=h]
    \textbf{User Prompt: Motif}
    \tcblower 
    \small
    Evaluate the consistency of [Motif] (the local geometric shape, waveform, or repeating structural unit). \\
    \textbf{Visual Evaluation Rubrics:} \\
    1. Treat the local pattern as a ``Shape Template''. \\
    2. STRICTLY IGNORE horizontal shifts (phase/timing offsets) and global vertical stretching (absolute amplitude). \\
    3. FOCUS ONLY on internal relative geometry: Check the peak-valley alternating rhythm, the relative height/width between adjacent peaks, and the sharpness/smoothness of the curves. \\
    4. [Exception for Amplitude Collapse]: This rule overrides Rule \#2 in extreme cases. If the forecast's amplitude is so severely compressed that its fluctuations become visually indistinguishable from random noise OR it fails to represent the essential structural differences between major peaks and minor valleys present in the ground truth, then the motif is considered broken. In such cases, rate as Severe Structural Mismatch (Score 2) or Complete Collapse (Score 1). \\
    \textbf{Scoring Rubric:} \\
    \textbf{5 (Perfect Replication):} The visual ``shape templates'' match exactly. Perfectly captures both the macro skeleton (peak-valley order) and micro-details (sharpness, local kinks). \\
    \textbf{4 (Structurally Intact with Minor Flaws):} The core visual skeleton is correct (the sequence of peaks and valleys matches perfectly). Allowed visual flaws: Slight smoothing (sharp peaks become rounded), minor top-clipping (flattened peaks), or slight distortions in relative proportions that do not change the recognizable shape. \\
    \textbf{3 (Noticeable Shape Degradation):} The basic up-and-down fluctuation exists, but the internal template is broken. Visual cues: Missing secondary peaks/valleys, severe distortion of internal relative proportions (e.g., a dominant peak becomes tiny), or the shape visibly degrades into simpler waves for 30\%-50\% of the timeline. \\
    \textbf{2 (Severe Structural Mismatch):} Visual cues: The predicted shape is completely wrong despite matching the duration; OR the rhythm/frequency is completely mismatched (e.g., predicting dense waves when GT is sparse); OR the shape degrades into generic noise for 50\%-80\% of the timeline. \\
    \textbf{1 (Complete Collapse):} No structured local pattern exists. The prediction visually collapses into a featureless flat line, a simple straight trend, or pure random static.
\end{tcolorbox}

\begin{tcolorbox}[float=h]
    \textbf{User Prompt: Phase}
    \tcblower 
    \small
    Evaluate the consistency of [Phase] (horizontal time alignment, frequency rhythm, and timing of events). \\
    \textbf{Visual Evaluation Rubrics:} \\
    1. Macro Rhythm First: Check the overall frequency (are the waves/events packed at the same density?). This is a prerequisite. \\
    2. Micro Timing Second: Check the horizontal (left-right) alignment of specific peaks, valleys, and sudden changes. \\
    3. STRICTLY IGNORE vertical height differences (amplitude) or overall vertical shifts. Focus purely on the X-axis alignment. \\
    \textbf{Scoring Rubric:} \\
    \textbf{5 (Perfect Alignment):} The macro frequency matches perfectly. All key peaks, valleys, and sharp turns align exactly on the vertical axis with no left/right shift (no lag or lead). \\
    \textbf{4 (Accurate Rhythm with Minor Flaws):} The overall frequency matches the ground truth. Most inflection points align precisely. Allowed visual flaws: May miss a few minor, non-key bumps, or exhibit very slight left/right timing shifts in less than 30\% of the timeline. \\
    \textbf{3 (Noticeable Phase Shift or Degradation):} The general frequency/rhythm is correct, but has noticeable visual flaws. Visual cues: (1) Systematic Shift: The predicted curve is visibly shifted left or right (overall lag/lead) but is NOT completely out-of-phase (i.e., peaks do not align with valleys); (2) Missing Details: Misses a significant number of minor bumps, though major peaks still align; (3) Local Degradation: Loses rhythm or is visibly misaligned in 30\%-50\% of the timeline. \\
    \textbf{2 (Severe Frequency Mismatch):} Visual cues: The macro rhythm is visibly wrong (e.g., predicted waves are much wider, narrower, or sparser than the ground truth). Even if a few peaks accidentally line up, it must be penalized if the overall wave density/frequency is incorrect. \\
    \textbf{1 (Complete Loss of Rhythm):} Visual cues: For $>$50\% of the timeline, the prediction completely loses its rhythmic or event-driven features, visually degrading into a flat line, a smooth monotonic trend, or pure random noise.
\end{tcolorbox}

\begin{tcolorbox}[float=h]
    \textbf{User Prompt: Magnitude}
    \tcblower 
    \small
    Evaluate the consistency of [Magnitude] (the relative vertical stretch/amplitude and the absolute vertical height/bias). \\
    \textbf{Visual Evaluation Rubrics:} \\
    1. The ``Lowest Score'' Rule (CRITICAL): You MUST evaluate two sub-dimensions: (A) Relative Amplitude (the peak-to-valley vertical stretch) and (B) Absolute Height (the overall vertical shift/gap between the curves). Your final score MUST be the LOWER of these two evaluations. \\
    2. Ignore Rare Outliers: Do not heavily penalize the model for missing 1 or 2 isolated, extreme spikes. Focus on the magnitude of the regular/bulk fluctuations. \\
    \textbf{Scoring Rubric:} \\
    \textbf{5 (Perfect Fit):} Both sub-dimensions are nearly perfect. (1) Amplitude: The peak-to-valley vertical stretch matches perfectly. (2) Height: No visible vertical shift (any gap is $<$10\% of the average wave height). \\
    \textbf{4 (Good, Minor Flaws):} The lowest of the two sub-dimensions is at level 4. (1) Amplitude: Visibly similar in scale, but slightly compressed or expanded (captures $>$2/3 of the true stretch). (2) Height: Visible but small vertical shift. The vertical fluctuation ranges (bounding boxes) of the two curves heavily OVERLAP. \\
    \textbf{3 (Severe Smoothing OR Severe Shift):} The lowest of the two sub-dimensions is at level 3. (1) Amplitude (Over-smoothed): Severely compressed. The predicted waves look noticeably flatter (captures approximately 1/3 to 2/3 of the true stretch). (2) Height (Severe Shift): Massive vertical gap. The bottom valleys of one curve barely touch the middle/top of the other (very little overlap in their vertical ranges). (3) Local Degradation: Amplitude dies out or shifts drastically in 30\%-50\% of the timeline. \\
    \textbf{2 (Extreme Smoothing, but still some visible fluctuation):} The lowest of the two sub-dimensions is at level 2. (1) Amplitude (Extreme Smoothing): The ground truth has massive waves, and the prediction shows some visible, albeit very small, fluctuations. The predicted amplitude is clearly present but captures significantly less than 40\% of the true stretch. (2) Height (Extreme Disconnect): The two curves are completely separated vertically with ZERO OVERLAP in their Y-axis ranges. (3) Severe Degradation: Amplitude dies out to a flat line in 50\%-80\% of the timeline. \\
    \textbf{1 (Collapse or Hallucination):} Visual cues: The forecast's fluctuations are so minor that they appear as `texture' or `noise' rather than distinct, individual waves. If you look at the forecast line and cannot clearly identify separate, meaningful up-and-down wave patterns that mirror the \textit{intent} of the ground truth, it is a collapse. OR, Amplitude (Near Collapse): The prediction shows almost no vertical stretch, appearing as a nearly flat line, even if it follows the general trend.
\end{tcolorbox}

\subsubsection{Macro-level Evaluation}

\begin{tcolorbox}[float=h]
    \textbf{User Prompt: w/o knowledge}
    \tcblower 
    \small
    You are a senior data science reviewer. You have been provided with a scoring report for a time series forecast across four fundamental dimensions. Your task is to provide a final, holistic ``General'' score (1-5) based on this report and the provided chart. \\
    \textbf{Atomic Dimension Scoring Report:} \\
    - Trend (Macro Trajectory) Score: \{trend\_score\} \\
    - Motif (Local Shape) Score: \{motif\_score\} \\
    - Phase (Rhythm \& Timing) Score: \{phase\_score\} \\
    - Magnitude (Amplitude \& Level) Score: \{magnitude\_score\} \\
    \textbf{Dimension Meanings for Reference:} \\
    - \textbf{Trend:} Governs the long-term trajectory and major turning points. \\
    - \textbf{Motif:} Defines the local geometric shape and waveform patterns. \\
    - \textbf{Phase:} Controls the temporal alignment, rhythm, and timing.  \\
    - \textbf{Magnitude:} Dictates the vertical amplitude and absolute level. \\
    \textbf{Scoring Rubric for [General Visual Fidelity]:} \\
    \textbf{5 (Excellent / Indistinguishable):} The prediction is visually almost identical to the ground truth. Perfectly captures the complete structural and dynamic essence, appearing as a near-perfect replica. \\
    \textbf{4 (Good / Highly Similar):} The prediction is clearly a ``twin'' of the ground truth. It successfully reproduces all core dynamics and major structural features, with only minor, non-misleading deviations in local details. \\
    \textbf{3 (Acceptable / Skeleton Preserved):} The prediction retains the primary structural skeleton (e.g., the main trend and rhythm) but suffers from a noticeable loss of detail or contains significant, obvious errors in at least one dimension. \\
    \textbf{2 (Poor / Structurally Distorted):} The prediction is fundamentally dissimilar to the ground truth. It presents a structurally distorted or misleading representation, failing to capture key dynamics (e.g., major trend direction, core frequency). \\
    \textbf{1 (Failure / No Correlation):} The prediction shows no discernible correlation with the ground truth. It has visually degenerated into a meaningless flat line, a simple trend, or a hallucinatory pattern.
\end{tcolorbox}

\begin{tcolorbox}[float=h]
    \textbf{User Prompt: Covid Death}
    \tcblower 
    \small
    Evaluate the prediction of [COVID-19 Cumulative Deaths]. \\
    \textbf{Evaluation Criteria:} \\
    1. \textbf{Monotonicity (Physical Constraint):} Cumulative data should be non-decreasing. Differentiate between a macro-level failure (sustained downward trend) and a micro-level flaw (local zigzags between adjacent points). \\
    2. \textbf{Error Direction (Risk Assessment):} Overestimation (prediction > ground truth) acts as a safety buffer for medical resources. Underestimation (prediction < ground truth) poses severe risks of resource shortages. \\
    3. \textbf{Error Magnitude (Fit Quality):} Assess whether the gap between the prediction and ground truth is negligible, stable (bounded), or diverging. \\
    \textbf{Scoring Rubric:} \\
    \textbf{5 (Excellent / Near-Perfect Fit):}
Monotonicity: Strictly non-decreasing.Deviation: Visually indistinguishable from the ground truth. Any deviations are marginal, non-compounding, and do not alter the overall trajectory. \\
    \textbf{4 (Good / Safe Overestimation):}
Monotonicity: Strictly non-decreasing.
Deviation: Consistently overestimates (prediction > ground truth), but the vertical gap remains stable and bounded. The predicted curve is structurally parallel to the ground truth, representing a safe and reasonable resource buffer. \\
    \textbf{3 (Fair / Sub-optimal Fit):}
Monotonicity: Generally non-decreasing. (Strictly NO visible downward drops; flat segments are acceptable).
Deviation: Falls into one of two specific cases: \\
(a) Extreme Overestimation: The positive gap is excessively large or exponentially diverging, indicating unrealistic resource allocation. \\
(b) Mild Underestimation: The prediction is slightly below the ground truth but strictly preserves the correct trend shape without diverging further. \\
    \textbf{2 (Poor / Severe Underestimation or Noisy Output):} \\
(a) Severe Underestimation: Consistently and significantly underestimates (prediction < ground truth), leading to high-risk resource shortages. \\
(b) Local Non-monotonicity: The overall macro trend is flat or upward, but the curve contains visible local downward fluctuations (zigzags/dips) between adjacent points, showing a lack of physical constraints. \\
    \textbf{1 (Failure / Macro Physical Violation):}
Monotonicity: The predicted curve exhibits a sustained downward slope or a gradual negative drift across the forecast horizon (e.g., the end point is visibly lower than the start point). The model completely fails to learn the fundamental macro-level constraint that cumulative deaths cannot decrease.
\end{tcolorbox}

\begin{tcolorbox}[float=h]
    \textbf{User Prompt: ECG}
    \tcblower 
    \small
    Evaluate the prediction of [ECG (Electrocardiogram) Signals]. \\
    \textbf{Visual Evaluation Rubrics:} \\
    1. \textbf{Count the Spikes (Rhythm):} Count the most prominent, sharpest peaks (QRS complexes) in both the ground truth and the prediction. Are the total counts exactly the same? \\
    2. \textbf{Check the Sharpness (Morphology):} Look at the predicted spikes. Are they sharp and needle-like (V-shaped/inverted V) or rounded and blunt (U-shaped/sine-wave)? \\
    3. \textbf{Check the Alignment and Baseline:} Do the spikes align horizontally? Is the baseline stable or drifting? \\
    \textbf{Scoring Rubric:} \\
    \textbf{5 (Perfect Diagnosis):}
Visual cues: The prediction perfectly replicates the rhythm and the detailed P-QRS-T morphology. The peaks align perfectly on the time axis, and the sharpness of the spikes matches the ground truth exactly. \\
    \textbf{4 (Clinically Safe - Minor Flaws):}
Visual cues: The total count of the main spikes matches exactly (correct heart rate). It accurately captures the rhythm. It may have minor baseline drift, slight vertical amplitude errors, or very small left/right phase shifts, but these do not affect the clinical heartbeat count. \\
    \textbf{3 (Diagnostic Ambiguity - Morphological Degradation):} 
Visual cues: The total count of the main spikes is EXACTLY the same as the ground truth, BUT the local shapes are visibly degraded. The sharp QRS spikes become noticeably wider or blunter, or the smaller secondary waves (P/T waves) are completely lost, creating a risk of misdiagnosis. \\
    \textbf{2 (High-Risk Misleading - Rhythm or Polarity Error):}
Visual cues: Fatal clinical errors. The model predicts MORE or FEWER main spikes than the ground truth (rhythm/heart rate error). OR, the peaks are flipped upside down (polarity reversal). OR, the sharp spikes are extremely smoothed out into gentle, rolling sine waves. \\
    \textbf{1 (Loss of Biological Meaning):}
Visual cues: The prediction completely loses the key fluctuations of the true signal. It degenerates into a flat horizontal line, a simple monotonic trend, pure random noise, or waves that have absolutely no structural match to the ground truth. \\
\end{tcolorbox}

\begin{tcolorbox}[float=h]

    \textbf{User Prompt: Traffic Speed}

    \tcblower 

    \small

    Evaluate the prediction of [Traffic Speed]. 

    \textbf{Visual Evaluation Rules:} 

    1. \textbf{Identify the "Valleys":} In traffic speed, normal flow is a high flat line, and congestion is a sudden downward drop forming a "deep valley".

    2. \textbf{Check Timing (Left vs. Right Shift):} Focus on the exact moment the curve starts to drop. Is the predicted drop shifted LEFT (early warning - highly preferred) or RIGHT (lagging/after-the-fact - strictly penalized)? 

    3. \textbf{Ignore Minor Vertical Gaps:} Tolerate absolute vertical distance errors during the high flat (free-flow) periods. Focus primarily on the shape, timing, and existence of the valleys. 

    \textbf{Scoring Rubric:} 

    \textbf{5 (Perfect Scheduling):}

Visual cues: The start (drop), bottom, and recovery of the congestion "valley" align perfectly on the time axis. The shape and timing are a perfect match.

    \textbf{4 (Early Warning / Safe Redundancy):}

Visual cues: The predicted valley starts to the LEFT of the ground truth (early warning). OR, the predicted valley is deeper than the ground truth (safely overestimating congestion). Minor vertical gaps during the high flat periods are completely acceptable.

    \textbf{3 (Reactive / Friction Cost):}

Visual cues: The predicted drop starts slightly to the RIGHT of the ground truth (minor lag). OR, the prediction shows a medium-sized valley while the ground truth is a high flat line (false alarm). OR, the prediction captures the valley but it is visibly much shallower than the true deep valley (severe underestimation of congestion).

    \textbf{2 (Severe Misleading / Missed Opportunity):}

Visual cues: The predicted drop is shifted severely to the RIGHT (massive lag, starting to drop only when the true congestion is already ending). OR, the ground truth shows a massive deep valley, but the prediction only shows a microscopic, barely visible dip (missing a major jam).

    \textbf{1 (Catastrophic Failure):}

Visual cues: Complete visual disconnect. The ground truth shows a severe "deep valley" (massive congestion), but the prediction remains a high, stable flat line (predicting free flow). OR vice versa (hallucinating a massive jam on a completely clear road). OR missing ANY severe deep valley.

\end{tcolorbox}

\end{document}